\documentclass[nonblindrev]{informs3}  
\DoubleSpacedXI

\usepackage{graphicx}
\usepackage{subfig, epsfig}
\usepackage{natbib}
\usepackage{makecell}
\usepackage{amsmath}
\usepackage{booktabs}
\usepackage{multirow}
\usepackage{bm}
 \bibpunct[, ]{(}{)}{,}{a}{}{,}%
 \def\bibfont{\small}%
\TheoremsNumberedThrough     
\ECRepeatTheorems
\EquationsNumberedThrough  

\begin{document}



\RUNTITLE{Improving Accuracy Without Losing Interpretability:  A ML Approach for Time Series Forecasting}
\vspace{2cm}

\TITLE{Improving Accuracy Without Losing Interpretability:  A Machine Learning Approach for Time Series Forecasting}

\ARTICLEAUTHORS{Yiqi Sun\\ \textit{Department of Industrial Engineering, Tsinghua University, Beijing, China}\\ Zhengxin Shi, Jianshen Zhang, Yongzhi Qi, Hao Hu \\ \textit{Intelligent Smart Y, JD.com, Beijing, China} \\   Zuojun Max Shen \\ \textit{College of Engineering, University of
California, Berkeley, Berkeley, California, USA} }

\ABSTRACT{In time series forecasting, decomposition-based algorithms break aggregate data into meaningful components and are therefore appreciated for their particular advantages in interpretability. Recent algorithms often combine machine learning (hereafter ML) methodology with decomposition to improve prediction accuracy. However, incorporating ML is generally considered to sacrifice interpretability inevitably. In addition, existing hybrid algorithms usually rely on theoretical models with statistical assumptions and focus only on the accuracy of aggregate predictions, and thus suffer from accuracy problems, especially in component estimates. In response to the above issues, this research explores the possibility of improving accuracy without losing interpretability in time series forecasting.  We first quantitatively define interpretability for data-driven forecasts and systematically review the existing forecasting algorithms from the perspective of interpretability. Accordingly, we propose the W-R algorithm, a hybrid algorithm that combines decomposition and ML from a novel perspective. Specifically, the W-R algorithm replaces the standard additive combination function with a weighted variant and uses ML to modify the estimates of all components simultaneously. We mathematically analyze the theoretical basis of the algorithm and validate its performance through extensive numerical experiments. In general, the W-R algorithm outperforms all decomposition-based and ML benchmarks. Based on P50\underline{ }QL, a common evaluation indicator for quantile prediction, the algorithm relatively improves by $8.76\%$ in accuracy on the practical sales forecasts of JD.com and $77.99\%$ on a public dataset of electricity loads.
 This research offers an innovative perspective to combine the statistical and ML algorithms, and JD.com has implemented the W-R algorithm to make accurate sales predictions and guide its marketing activities.

}
\KEYWORDS{time series forecasting,   time series decomposition, machine learning,  interpretability} 

\maketitle


%


\vspace{-0.4cm}

\section{Introduction}
\label{sec:intro}

An algorithm with high accuracy, interpretability, and ability to deal with complex empirical data is the ultimate goal of forecasting.
In time series forecasting, researchers have developed various approaches for different scenarios with distinct superiorities. Classical statistical methods, such as exponential smoothing (ETS for short) and autoregressive integrated moving average (ARIMA for short) models, have clear mathematical structure but can only handle simple time series data \citep{Gooijer2006}. These algorithms cannot capture the influence of exogenous variables and are, therefore, not satisfactory in accuracy, especially for non-stationary time series. The subsequent algorithms split into two main streams.
One stream retains the statistical structure, focuses more on interpretability, and plays a crucial role in the industry. Typical examples are the state-space models that simplify the variation of time series through a discrete-time stochastic model \citep{Tep1985, Snyder2017}, and decomposition-based methods that split the aggregated data into interpretable components \citep{Hyndman2021}, such as the STL and Prophet algorithms \citep{Cleveland1990, Sean2018}.
These algorithms either introduce the exogenous variables structurally or complicate the computational procedure for a better fit, but the improvement in prediction accuracy is limited.
On the other hand, another stream prefers extreme prediction accuracy through black-box ML models. Related algorithms pay more attention to the efficient use of extensive empirical data and nonlinear relationships between the data and predictions. The majority of this stream is the neural networks (hereafter NN), such as Support Vector Machines \citep[e.g.,][]{KIM2003}, Long-Short Term Memories \citep[][]{LSTM1997}, and Gate Recurrent Units \citep{GRU2014}. Nevertheless, the lack of interpretability has become a critical obstacle for these algorithms in practice \citep{molnar2022}.


Researchers have turned to hybrid algorithms that combine classical statistical models with black-box ML algorithms to improve forecasting accuracy and data processing ability.
Most studies, including ours, focus on time series decomposition for its specific advantages in interpretability and flexibility. Additionally, proper decomposition extracts underlying patterns from aggregate panel data \citep{Hyndman2021} and provides valuable insights for operations and supply chain management \citep{STR2021}.
Based on the popular Prophet algorithm, \citet{Triebe2021} proposes NeuralProphet, which uses statistical methods to predict the trend and seasonality components and NN for the auto-regression and exogenous effects.
Following the statistical decomposition procedure, both Prophet and NeuralProphet predict the individual components through independent modules. In contrast, the N-BEATS and NBEATSx algorithms go further and deviate significantly from the traditional statistical approaches. Compared to the NeuralProphet, these algorithms output the estimated components simultaneously through a connected NN consisting of different modules for the corresponding components and an additional module to predict the residuals between the simple addition and target value \citep{Oreshkin2019, OLIVARES2022}. For interpretability considerations, these algorithms, including the classic and hybrid methods, usually generate the final aggregate prediction by simply adding the components.

While claimed with multiple benefits, several remaining problems seriously restrict the further development of hybrid algorithms.
First, the contradiction between interpretability and accuracy is likely inevitable \citep{Murdoch2019}. Therefore, with improved accuracy, hybrid algorithms are usually considered less interpretable after incorporating ML. We attribute this perception to the lack of quantification in the existing definition framework.
To our knowledge, the existing definitions of ML interpretability are all ambiguous and abstract, such as those in \citet{Miller2017}, \citet{Murdoch2019}, and \citet{molnar2022}. In the absence of a well-defined concept, comparing the algorithms in interpretability usually relies on the model complexity \citep[]{Zhou2018}, which is improper in many cases. For example, the tree-based algorithms are often considered more interpretable than NN \citep[e.g., in][]{Guo2022}. However, when the depth of trees reaches a milestone, such as a hundred, the relative advantage in interpretability becomes negligible.
A more intuitive example is that a feed-forward NN with a single layer of 64 units is considered less interpretable than that of 32 units; however, they are about the same as people understand neither.

Moreover, the hybrid algorithms also succeed to the drawbacks of classical structural algorithms, which limit the further improvement in accuracy. In time series decomposition, the components are often generated through individual modules, and the final predictions are the simple addition of component estimates. These settings fail to capture the correlations among components and imply an assumption of independence. While it holds in some cases, the independence essentially relies on the components' mathematical definitions. For example, the trend and seasonality components are independent of each other as they respectively refer to the stable and periodic constituents of the time series. Accordingly, the ML modules have to strictly follow the structure of corresponding theoretical models, which restricts the potential for accuracy and efficiency. Worse yet, other components may be closely correlated, such as the impacts of promotion and advertising on sales.
The forecasting models' performance would be compromised when the assumption of independence is not satisfied, which is likely to occur in reality.

In addition to the mandatory assumptions restricting the overall performance, there is a remaining issue for all decomposition-based algorithms: the less accurate prediction of components. As only aggregate data are available, isolating the effects of various factors is tricky, and decomposition quality is difficult to examine. The measures breaking the components in existing algorithms usually rely on manual settings. For example, the parameters dividing the trend and seasonality components, which are supposed to change as appropriate automatically, are often pre-determined manually \citep[e.g., in][]{Cleveland1990, STR2021}.
For the influence of exogenous variables, the estimates are usually simple linear impacts that are incomplete \citep[e.g., in][]{Triebe2021, OLIVARES2022}. These problems lead to inaccurate component predictions, negatively affecting the total prediction accuracy and challenging the value of induced insights from proper decomposition.

To summarize, we focus on the following problems in this study and propose corresponding solutions.

\textbf{\textit{1. How to properly define interpretability in a quantitative manner?}}

As noted before, a proper definition of interpretability is critical to the further development of hybrid algorithms. In this work, we propose an innovative definition of interpretability for data-driven forecasts, which divides the general concept into input and function interpretability. The novel definition allows us to compare the interpretability among algorithms quantitatively.
Accordingly, we systematically review the existing algorithms from the perspective of interpretability, which reveals the possibility of improving accuracy without losing interpretability.

\textbf{\textit{2. Based on the quantitative definition of interpretability, how to improve the accuracy of the decomposition-based algorithms without losing interpretability?}}

Inspired by the innovative definition of interpretability, we propose the W-R algorithm, which utilizes a weighted function to combine the components. Specifically, in the W-R algorithm,
$$\textit{Final Prediction} = \sum_{i}\left( \textit{Weight} \times \textit{Preliminary Prediction of Component $i$} \right)+ \textit{Residual,}$$
where the weights and residuals are given by ML modules automatically. Note that the weighted combination function is equivalent to simple addition in interpretability as the parts multiplying weights can be considered as new estimates of components.
We validate the advantages of the W-R algorithm through theoretical analysis and extensive numerical experiments.

\textbf{\textit{3. How to improve the prediction accuracy of components only with aggregate data? }}

In addition to the overall performance, we are also curious about the forecasting accuracy of components, which is inherently tricky to analyze. Therefore, we first focus on a specific case with two components and derive the mathematical solutions and properties. For the special case, we derive conditions for both components to improve under the weighted mechanism,  finding that the conditions are likely to hold when the weights are in a moderate interval around $1$. We then promote the findings into conjectures for the more general case and validate them through numerical experiments. To our knowledge, we are the first to consider the accuracy of components inside time series decomposition from a practical perspective.

\textbf{\textit{4. Is the conflict between interpretability and accuracy always inevitable?}}

Taking the solutions to previous research problems together, we believe that the conflict between interpretability and accuracy is not always inevitable as long as we consider interpretability from a more general perspective. In the W-R algorithm, we implement the weighted combination function, which is different in complexity from the simple addition, but of the same level from the perspective of human understanding. Therefore, the W-R algorithm improves multi-dimensional accuracy without loss of interpretability.

The remainder of the paper is organized as follows.
In Section \ref{sec:lit}, we present the quantitative definition of interpretability and systematically review the related algorithms. Section \ref{sec:model} introduces the W-R algorithm with the weighted combination function, including theoretical analysis and concrete models for implementation. Section \ref{sec:result} reports the numerical results on the private and public datasets, and Section \ref{sec:conclusion} concludes the paper. The E-companions (online supplements) present the mathematical proofs and the details for numerical experiments. In addition, we also provide background information on related algorithms for the convenience of readers.

\section{Definition and Literature}
\label{sec:lit}
\vspace{0.2cm}

\subsection{Interpretability in Data-driven Forecasting}
\label{subsec:ldefinition}

Let's start with a fundamental but persistent unanswered question: What is interpretability for forecasting algorithms? Existing definitions in literature are usually broad qualitative descriptions \citep{Bertsimas20191,molnar2022}. For example, \citet{Miller2017} defines interpretability as the extent to which humans can understand the reasons for outputs from ML systems. And in \citet{Murdoch2019}, interpretability refers to the ability to extract domain relationships contained in the data or learned by the model.
Due to the stupefying complexity within, researchers have decomposed the concept from various perspectives for clarity.
\citet{du2019} differentiates ML interpretability into global and local interpretability, depending on whether one can understand all about the model or just a few results. Similarly, \citet{molnar2022} divides interpretable ML techniques into intrinsic interpretability, which arises from the ML algorithm itself, and post-hoc interpretability, which is model-agnostic and can be widely applied to all existing algorithms.

\begin{figure}[!htbp]
		\centering
	\caption{The General Procedure of Data-driven  Forecasting}
\vspace{0.2cm}
 \includegraphics[width=4in]{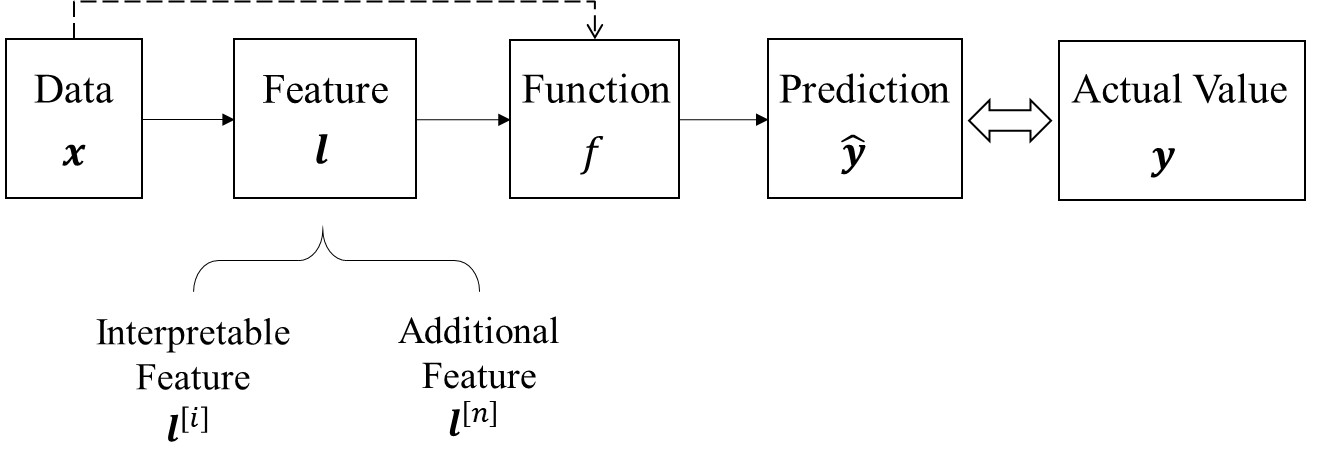}		\label{fig:framework0}
\vspace{-0.6cm}
\end{figure}

Figure \ref{fig:framework0} presents the general procedure of data-driven forecasting. There are four key elements inside, including the original data $\bm{x}$, the extracted features $\bm{l}$, the output prediction $\bm{\hat{y}}$, and the function $f$ capturing the relationship between the inputs and outputs. Note that the labels in bold represent vectors or matrices.
While the model may include both the data and features as inputs in practice, we only consider the features as inputs, i.e.,
\begin{eqnarray}
\bm{\hat{y}}=f(\bm{l}) \label{eq:base}
\end{eqnarray}
 mathematically, as the data can be considered as features without any additional processing.

\begin{figure}[!htbp]
		\centering
	\caption{The Quantitative Definition Framework of Interpretability}
\vspace{0.2cm}
 \includegraphics[width=4.5in]{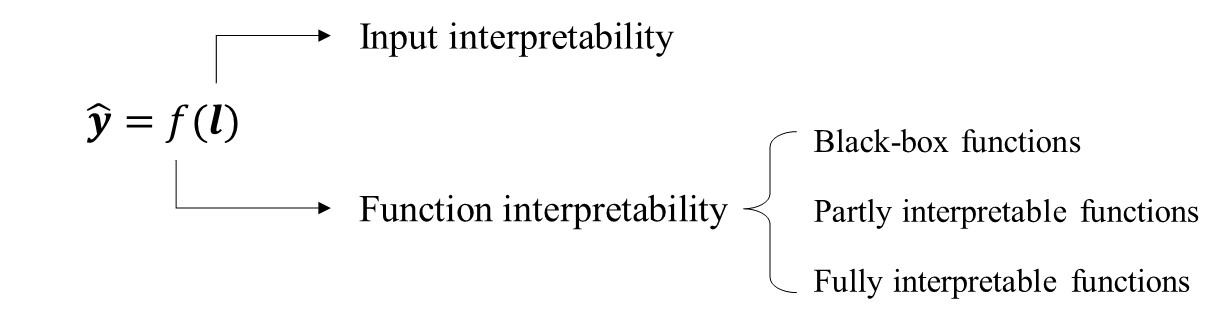}		\label{fig:interpretability}
\vspace{-0.3cm}
\end{figure}

According to Eq.~(\ref{eq:base}), we divide the general interpretability of data-driven forecasting into the \textit{input} and \textit{function interpretability}, and illustrate the structure in Figure \ref{fig:interpretability}. The level of \textit{input interpretability} refers to the relative proportion of interpretable features in inputs. Specifically, the interpretable features $\bm{l}^{[i]}$ refer to those with specific meanings and can be understood by people directly, while other additional features $\bm{l}^{[n]}$ only serve to improve the forecasting accuracy. In time series forecasting, typical interpretable features include the moving average features, such as the average sales in the past 7/14/28 days, and the features of exogenous variables, e.g., if there's a festival. Generally speaking, input interpretability is the basis of interpretable forecasting.

In contrast, \textit{function interpretability} refers to the  interpretability of function $f$, and is the primary deficiency for most  less interpretable ML algorithms. From this perspective, we categorize the functions in existing algorithms into fully interpretable,  partly interpretable, and black-box functions,  denoted by $f_i$, $f_p$, and $f_n$, respectively. 
The fully interpretable functions have simple structural expressions that people can understand directly, such as linear regression and a single decision tree with limited depth.
Meanwhile, partly interpretable functions are those with complicated mathematical expressions that cannot be interpreted directly. Corresponding algorithms usually have identified rules for model generation, such as the random forests, the XGBoost, and other tree-based algorithms \citep[see][for detail]{Breiman2001,Chen2016}. In other words, people can directly understand the individual steps in these algorithms but not the aggregated outputs.
 Finally, the black-box functions refer to those that cannot be expressed by closed-form mathematics. The corresponding algorithms are usually not interpretable in both the generation procedure and output predictions. Representative examples in this category are the NN algorithms, including the popular Convolutional Neural Networks (CNNs), Recurrent Neural Networks (RNNs), and hundreds of variants. For the convenience of understanding, we list the key differences among these functions in Table \ref{tab:definition}.

\begin{table}[!htbp]
\centering
\footnotesize
\setlength\tabcolsep{4pt}
\renewcommand\arraystretch{0.8}
\caption{Classifications of Functions in Interpretability}
\begin{tabular}{lllllll}
\hline\\[-0.4cm]
    \multirow{4}{2cm}{\textit{Category of Functions}}                               & \multicolumn{2}{c}{\textbf{For the function}}                                                                                    & \multicolumn{2}{c}{\textbf{For corresponding algorithm}}                                                                   \\
\\[-0.4cm]\cmidrule(r){2-5}\\[-0.7cm]
                                  &\makecell[l]{Does
  the function \\[-0.3cm] have mathematical \\[-0.3cm] expression?}   & \makecell[l]{Does
  the mathematical \\[-0.3cm] expression simple enough \\[-0.3cm] to understand directly?}   & \makecell[l]{Can we separate \\[-0.3cm]
  the algorithm  into \\[-0.3cm]interpretable steps?}   & \makecell[c]{Typical examples } \\[-0.3cm]
\\[-0.2cm]\hline\\[-0.4cm]
\textit{Fully
  interpretable}   & \multicolumn{1}{c}{Yes}                            & \multicolumn{1}{c}{Yes}                                                     & \multicolumn{1}{c}{Yes}                                    & \multicolumn{1}{c}{Structural
  models}                    \\ \\[-0.4cm]
\textit{Partly
  interpretable} & \multicolumn{1}{c}{Yes}                            & \multicolumn{1}{c}{No}                                                      & \multicolumn{1}{c}{Yes}                                    & \multicolumn{1}{c}{Rule-based
  algorithms}            \\ \\[-0.4cm]
\textit{Black-box}          & \multicolumn{1}{c}{No}                             & \multicolumn{1}{c}{No}                                                      & \multicolumn{1}{c}{No}                                     & \multicolumn{1}{c}{Neural
  networks}\\ \\[-0.4cm]
\hline
\end{tabular}
\label{tab:definition}
\end{table}
\vspace{-0.2cm}

In data-driven forecasting, an algorithm is interpretable if and only if the inputs and function are all interpretable. Mathematically, $\bm{\hat{y}}=f(\bm{l})$ is interpretable when $\bm{l}=\bm{l}^{[i]}$ and $f(\cdot)=f_i(\cdot)$. In other words, we should utilize ML modules to generate interpretable features and functions to improve accuracy without sacrificing interpretability. Note that our definition of interpretability mainly focuses on the interpretation of results rather than the detailed prediction procedures; therefore,
there is still room for using black-box modules inside the algorithms. For example, as mentioned before, recent time series decomposition algorithms often utilize ML algorithms to predict the interpretable features \citep[e.g., in][]{Oreshkin2019,Triebe2021}. The final outputs in these algorithms are the addition of the components, which is functionally interpretable.

\subsection{Literature Review}

\begin{figure}[!htbp]
		\centering
	\caption{Classification Framework for Algorithms in Data-driven Forecasting}
\vspace{0.2cm}
 \includegraphics[width=6.5in]{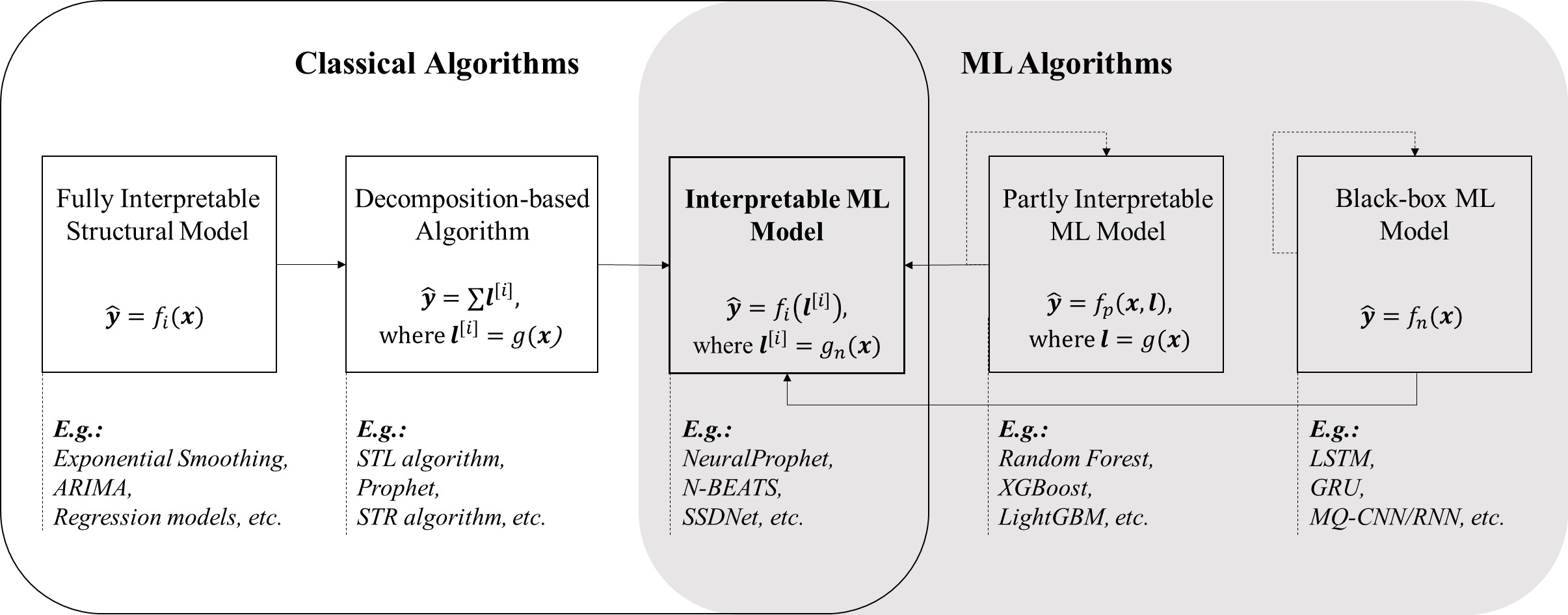}		\label{fig:model_int}
\vspace{-0.8cm}
\end{figure}

We then systematically review the algorithms in time series forecasting from the perspective of interpretability.
Figure \ref{fig:model_int} presents a unifying framework to classify related algorithms. There are mainly two streams of algorithms in data-driven forecasting, classical algorithms with high interpretability but low accuracy and ML algorithms that are exactly the opposite.

\subsubsection{Classical Algorithms with Superior Interpretability}
\label{subsec:classical}

As mentioned in Section \ref{sec:intro}, the early methods in time series forecasting are usually simple statistical/structural models with full interpretability. Typical examples include the ETS methods proposed by \citet{Holt1957} and \citet{Winters1960}, ARIMA from \citet{Box1970}, and the historic regression models \citep[e.g.,][]{Hyndman2021}. These algorithms usually work directly on original time series data and have clear mathematical expressions for both the calculation procedure and output predictions. However, as we may expect, most of these algorithms cannot handle complex practical data and therefore are less satisfying in accuracy.

Subsequent researchers draw on the idea from \citet{Holt1957} and \citet{Winters1960} and generate a series of decomposition-based algorithms to improve forecasting accuracy. Up to now, related popular methods include the STL, Theta, Prophet, STR and STD algorithms for general time series \citep{Cleveland1990,Assimakopoulos2000,Sean2018,STR2021,STD2022}. In general, these algorithms first calculate the components through individual modules and then combine them additively \citep{Gooijer2006,Hyndman2021}. Common components consist of the trend, seasonality, and impacts of various exogenous variables (e.g., weather and holiday). Note that the components here are essentially equal to the features $\bm{l}$ in Figure \ref{fig:framework0}, and are likely to be interpretable to ensure the overall interpretability.
Meanwhile, the modules calculating the components may be intricate to improve forecasting accuracy. For instance, the components in the STL algorithm, including the seasonality and trend, are generated by \textit{Loess}, a weighted regression method based on the K-nearest-neighbor algorithm. In Prophet, the trend component is captured by a logistic growth model and a piecewise linear model, while the seasonality is described by the Fourier series.

Similar to the fully-interpretable structural models, the decomposition-based algorithms maintain distinctive superiority in interpretability and therefore play critical roles in forecasting practice \citep[e.g., ][]{Snyder2017,Chun2021}. However, most methods are less efficient in processing the massive data generated by daily operations, resulting in accuracy issues. Moreover, as mentioned in Section \ref{sec:intro}, these algorithms usually rely on underlying assumptions, such as the pre-determined structure of components and the independence among components. These assumptions restrict the scope of application of the algorithms. Worse yet, the premises are likely to be broken in reality, making the performance less satisfying.

\subsubsection{ML Algorithms with High Accuracy}

Contrary to classical methods, ML algorithms can handle complex data and are superior in accuracy but less satisfying in interpretability. A majority of the ML algorithms are the black-box NN algorithms, including the multilayer perceptron \citep[MLP for short, e.g.,][]{Nesreen2010}, CNN \citep[e.g.,][]{Borovykh2019}, RNN \citep[see][]{Hansika2021}, and attention-based transformers \citep{attention2017}. Note that while the transformers perform better in interpretability than CNN and RNN, the superiority is essentially not much from our perspective. The core of transformers is the self-attention function, which captures the self-interrelations of the outputs with a precise mathematical expression. However, the function generally serves as an intermediate step between other black-box modules in the application, making the whole algorithm less interpretable. Therefore, based on interpretability defined in Section \ref{subsec:ldefinition}, the pros and cons of attention-based algorithms are consistent with other NN models.

In time series forecasting, the popular NN models include the LSTM proposed by \citet{LSTM1997}, GRU from \citet{GRU2014}, and TCN from \citet{TCN2016}. Recent NN algorithms usually aim to meet specific requirements from practical applications, e.g., the WaveNet and DeepAR to predict quantiles \citep{WaveNet2016,DeepAR2020} and MQ-CNN/RNN for multiple quantile predictions with long forecasting horizon \citep{Wen2017}. These algorithms usually execute on normalized data and do not need additional assumptions on data or functions. However, while reported high accuracy, these algorithms are generally less interpretable and hence have problems in credibility, robustness, and troubleshooting \citep{molnar2022}.

Another category is the tree-based algorithms, with the highest interpretability among ML algorithms.
Early tree methods only include a single decision tree, such as the classification and regression trees proposed by \citet{CART1983}, therefore are fully interpretable when the number of nodes is limited. Later algorithms usually consist of several sub-trees for better robustness and accuracy, and the final output combines the outcomes from separate sub-trees. For example, the final prediction is the average of the sub-outputs in the random forecast algorithm \citep{Breiman2001}, and the addition in XGBoost and LightGBM \citep{Chen2016,LightGBM2017}.
As defined in Table \ref{tab:definition}, these algorithms are at least partly interpretable; However, as indicated in \citet{Murdoch2019}, there is a common conflict between interpretability and predictive accuracy. In these tree models, a large number of features are often necessary to ensure predictive accuracy, which decreases input interpretability. Moreover, the massive features also inevitably increase the number of nodes and subtrees in the model, and the high complexity leads to low interpretability for the function $f$. In practice, the tree-based algorithms are likely to be partly interpretable to ensure forecasting accuracy, except for some straightforward cases when the model only needs a single decision tree with several nodes to achieve satisfying predictions.

To summarize, there are mainly two streams of ML algorithms in time series forecasting, the black-box NN and the tree-based algorithms with partial interpretability. Generally speaking, ML models with more input features and more sophisticated structures perform better in accuracy, implying a conflict between interpretability and accuracy. As people continue to pursue high accuracy, the deficiency of interpretability has become a critical concern for applying ML in practice.

\subsubsection{Combination: Interpretable ML Model}

Researchers have integrated various methodologies to generate interpretable ML algorithms to simultaneously achieve nice accuracy and interpretability. These hybrid methods usually include two steps: (1) use ML to generate interpretable intermediate variables; (2) and then combine these intermediate outcomes through an interpretable function $f_i(\cdot)$. As indicated in Section \ref{sec:intro}, typical examples include the NeuralProphet, N-BEATS, NBEATSx, and SSDNet.
The NeuralProphet algorithm proposed by \citet{Triebe2021} succeeds to the structure of Prophet and generates the components through separate modules. Following Prophet, NeuralProphet uses statistical models in predicting the trend and seasonality but creatively utilizes NN to predict the effects of auto-regression and exogenous variables.
Based on simple feed-forward NN, the N-BEATS algorithm proposed by \citet{Oreshkin2019} uses sequential NN modules to simultaneously predict the components of trend, seasonality, and residuals.
Afterward, \citet{OLIVARES2022} considers the impacts of exogenous variables and extends N-BEATS to the NBEATSx. While the components are generated simultaneously, these two algorithms severely rely on the physical definitions of components. For specific components, the structure inside the corresponding module depends on the theoretical model, precisely, the polynomial regression for the trend and periodic functions (i.e., $sin(\cdot)$ and $cos(\cdot)$) for the seasonality part. Similarly, the SSDNet algorithm utilizes NN to simulate a state space model, another theoretical model capturing the time-related variations of the components \citep{SSDNet2021}. Consistent with other algorithms mentioned above, the SSDNet also divides the time series into components of seasonality, trend, and random error, and the final prediction is the simple addition of the components for specific forecasting periods.

These algorithms have several common characteristics and hence similar problems. First, all of the interpretable ML algorithms mentioned above only incorporate ML in the prediction of components. As a result, the deficiencies of the classic decomposition algorithms still hold for these algorithms.
The final predictions usually follow a simple additive form, i.e.,
\begin{eqnarray}
\bm{\hat{y}}=\sum_j l_{j}^{[i]}
\end{eqnarray}
where $l_{j}^{[i]}$ denotes the $j$th explainable component. The equation is simple with full interpretability; However, as discussed in Sections \ref{sec:intro} and \ref{subsec:classical}, without additional consideration on correlations, it implies the assumption of independence among components, and restricts the design of decomposition-based algorithms. Most algorithms only consider independent components, such as seasonality, trend, and residuals. The corresponding NN modules are designed following the deep logic of the theoretical models built on the  mathematical definitions. Other components, such as the effects of exogenous variables in NeuralProphet and NBEATSx, are usually modeled by a simple linear function without cross-component interrelations. That is, the ML modules only serve as the agent of theoretical models. Therefore, the benefits of ML, especially the tolerance of statistical assumptions, are reduced, which limits the forecasting accuracy and further application in practice.


Finally, it is worth mentioning that the problem of lacking independence is magnified when we turn to the specific time series on sales. Compared with general time series, sales are significantly influenced by several exogenous factors, which have been indicated in extensive empirical and theoretical literature. Taking these elements into consideration would help improve forecasting accuracy. For example, in addition to the components of seasonality and trend, the SCAN$*$PRO algorithm integrates price, competition, and advertising \citep{Wittink1988}, and the CHAN4CAST method further incorporates past sales, temperature, and holiday \citep{CHAN4CAST2005}. However, these factors are nearly impossible to be entirely independent of other factors. For example, price discounts and advertising always occur concurrently for the same promotion activities,  periodically or with intense competition. Meanwhile, the various influencing factors of a given seller may be highly consistent due to the same sales target or specific enterprise culture. In this case, the common additive combination function in existing algorithms is challenged from a fundamental perspective.

\subsubsection{Summary}

To summarize, incorporating ML in decomposition-based time series forecasting has become an inevitable trend for simultaneous benefits in accuracy and interpretability. ML methods have natural superiority in dealing with massive practical data and capturing non-linear relationships without strict statistical assumptions. However, existing relevant algorithms only utilize ML in the separate forecast of components, and the inappropriate assumption of independence still holds in the additive form of total predictions. The improper assumption restricts the accuracy of predictions, especially those on the decomposed components.

In this research, we propose W-R Algorithm, a hybrid forecasting algorithm following the interpretable ML stream but focusing more on the combination function $f_i(\cdot)$, in addition to the prediction of components. Specifically, the combination function follows a weighted linear form, where the parameters are generated from ML modules. Therefore, it is still interpretable but more accurate than the simple additive function. This idea is essentially similar to the optimal tree method proposed by \citet{Bertsimas2021}, which aims to improve a tree with proper design. However, our algorithm is built on a linear combination function rather than the trees; therefore, the decision spaces are of distinct forms. Moreover, while the optimal trees utilize mixed integer programming to determine the optimal splits, our algorithm uses ML to determine the improved parameters, proving efficient through theoretical analysis and numerical experiments.

\section{The W-R Algorithm for Interpretable Forecasting}
\label{sec:model}

This section details the W-R algorithm with the weighted combination function. Section \ref{subsec:structure} first introduces the algorithm's general structure, and Section \ref{sec:weight} theoretically analyzes the weighted mechanism. Finally, in Section \ref{subsec:concrete_models}, we describe the implementation details of the W-R algorithm.

\subsection{General Structure} 
\label{subsec:structure}
Consistent with the recent algorithms for time series forecast \citep[e.g.,][]{Wen2017,Lim2021}, our algorithm predicts the quantiles in multiple forecasting horizon. Let $N$, $t$, and $H$ respectively denote the number of components, time of forecasting, and forecasting horizon. Then the task is to predict the time series of the following $H$ periods from time $t$, denoted by $\bm{\hat{y}}=\{\hat{y}_{t+j}, \ \forall j\in\{1,\cdots, H\}\}$. The target is to minimize the difference between  $\bm{\hat{y}}$ and the actual value $\bm{y}$. 

\begin{figure}[!htbp]
		\centering
	\caption{General Structure of the W-R algorithm}

 \includegraphics[width=4.5in]{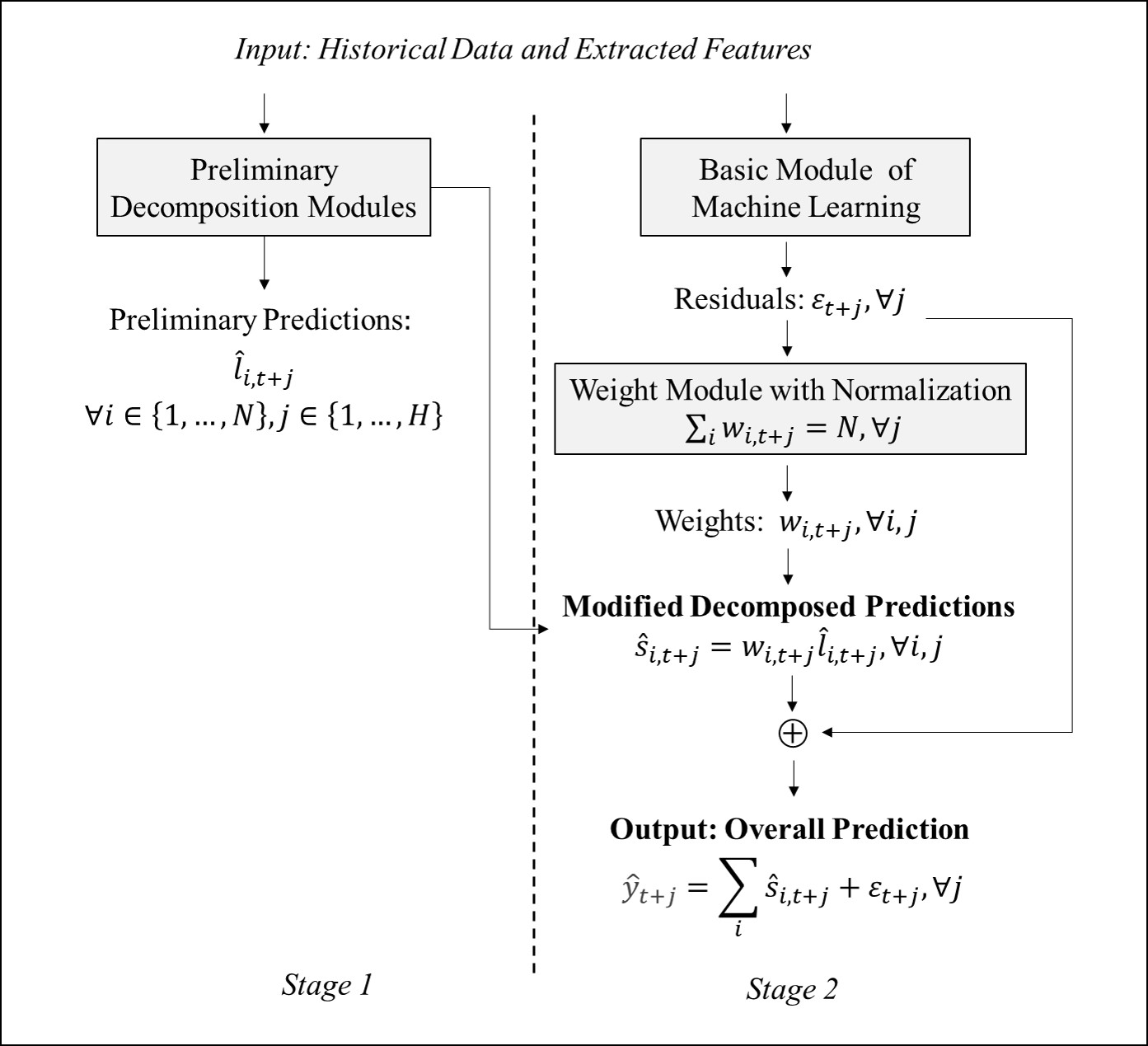}		\label{fig:framework}
\end{figure}

As illustrated in Figure \ref{fig:framework}, the W-R algorithm includes two connected stages.
Stage 1 generates the preliminary decomposition according to historical data and extracted features. Mathematically, the preliminary estimates of component $i$ for period $t+j$ is denoted by $\hat{l}_{i,t+j}$, where $i\in\{1,\cdots, N\}$, $j\in\{1,\cdots, H\}$. The specific approaches of preliminary decomposition are flexible, depending on the practical scenario and characteristics of the data.

Most existing decomposition-related algorithms for time series forecasting only include the above stage, in which the final predictions are the simple addition of the components \citep[e.g.,][]{Winters1960,Cleveland1990,Sean2018}. Through an additional ML module with a clear structure, the W-R algorithm further improves the preliminary estimates in stage 2. As highlighted in the algorithm's name, two key intermediate elements,   $\bm{w}$ and $\bm{\varepsilon}$, serve as the weights and residuals in the weighted combination function to modify the naive predictions.

Specifically, in stage 2, the original data and features are first put into a basic ML module and generate an intermediate variable $\bm{\varepsilon}=\{\varepsilon_{t+j}, \ \forall  j\in \{1,\cdots, H\}\}$. Compared with statistical methods, the ML algorithm better discovers the underlying non-linear relationship between the complex practical data and predictions.  The intermediate output $\bm{\varepsilon}$  is then put into a weight module, of which the output is denoted by $\bm{w}=\{w_{i,t+j}, \ \forall i\in \{1,\cdots, N\}, j\in \{1,\cdots, H\}\}$. Constrained by $\sum_iw_{i,t+j}=N$,  $\bm{w}$ serves as the weight factor to modify the decomposed predictions, and the modified predictions on components are $\hat{s}_{i,t+j}=w_{i,t+j}\hat{l}_{i,t+j},\ \forall i\in \{1,\cdots, N\}, j\in \{1,\cdots, H\}$. Finally, we merge the modified prediction  with $\bm{\varepsilon}$ by the  residual connection method \citep{RESNET2015}, hence the overall prediction is $\bm{\hat{y}}=\{\hat{y}_{t+j}=\sum_i \hat{s}_{i,t+j} +\varepsilon_{t+j}, \ \forall i\in \{1,\cdots, N\}, j\in \{1,\cdots, H\}\}$.  As it denotes the difference between $\hat{y}_{t+j}$ and the weighted sum of components, and we call $\varepsilon_{t+j}$ the \textit{residual}.  

The advantages of the W-R algorithm are manifold. First, the weighted combination function is a simple linear equation with full function interpretability. As a result, the algorithm is highly interpretable as long as the components are meaningful. Besides, the naive predictions on components are modified through the ML algorithm, hence are more accurate (we'll analyze this afterward) and valuable for subsequent analysis and decision-making. Moreover, through extensive numerical experiments, the W-R algorithm also performs better than the individual algorithms within, such as the basic ML module in stage 2. This phenomenon indicates the importance of domain knowledge; that is, information on proper decomposition would help model performance.
Finally, the W-R algorithm owns excellent flexibility in application. We can select different methods for preliminary decomposition and the basic ML module within to meet various requirements in practice.

\subsection{Theoretical Analysis on the Weight Mechanism}
\label{sec:weight}

We then explore the underlying mechanism behind the W-R algorithm theoretically. The core of the algorithm is the weighted combination function, i.e., 
\begin{eqnarray}
\hat{y}_{t+j}=\sum_{i=1}^{N} w_{i,t+j}\hat{l}_{i,t+j}+\varepsilon_{t+j}\ \ \textit{with}\  \ \sum_i w_{i,t+j}=N. \label{eq:function}
\end{eqnarray}
Originating from simple linear regression, the weighted mechanism is quite common in theory and practice. In our case, we allocate different weight parameters for different preliminary estimations (i.e., $w_{i,t+j}$ for $\hat{l}_{i,t+j}$), therefore our weighted mechanism is \textit{dynamic} rather than static as in regressions. This setting copes with the dynamic characteristics of estimation biases; that is, the estimation biases usually vary with different components and over time. As a result, the prediction accuracy would be better than the static model. Moreover, by adding weights to the individual components for distinct periods, the independence assumption is not required in our case. However, the incorporation of additional parameters significantly increases the dimension of the feasible region and the difficulty of solving. According to the following theoretical analysis, we take two measures in response.

As Eq.~(\ref{eq:function}) works for all forecasting horizons, we omit the subscript ${t+j}$ in the rest of this subsection for simplicity. As we expected, the residual term $\varepsilon$ helps improve the forecasting accuracy, as it accounts for the unconsidered components, such as the effects of some exogenous variables. As a result, we mainly want to discuss the effectiveness of the weight mechanism here, where the benchmark is the common additive function. Mathematically, the weighted combination function without residual is expressed as $\hat{y}=\sum_{i=1}^{N-1} w_{i} \hat{l}_i+(N-\sum_{i=1}^{N-1} w_{i})l_N$, while the additive combination function is $\hat{y}=\sum_{i=1}^N \hat{l}_i$. The actual relationship behind is supposed to be $y=\sum_{i=1}^{N-1} l_i$, where $l_i$ and $l_j$ can be correlated with each other when $i\neq j$.


\subsubsection{The Overwhelming Advantage in Overall Accuracy}
\label{sec:weight-1} 
It is not surprising that the weighted combination shows absolute superiority in forecasting accuracy compared with the simple addition.  For the convenience of mathematics, we first consider a special case with  $N=2$. We label the two components $l_1$ and $l_2$,  and the estimates are  $\hat{l}_1$ and $\hat{l}_2$, respectively. Then for the weighted combination function without residual, the prediction $\hat{y}= w  \hat{l}_1 +(2-w)\hat{l}_2$, with only 1 parameter $w$ to optimize. We only consider the case with unequal components, as   the prediction would stay constant  when $\hat{l}_1=\hat{l}_2$. Without loss of generality, we assume $\hat{l}_1> \hat{l}_2$.

\begin{proposition}
When $N=2$, there exists a unique weight parameter $w^*=\frac{{y}-2{\hat{l}}_2}{{\hat{l}}_1-{\hat{l}}_2}$ optimizing the weighted combination function without residual, and the weighted combination with optimized $w^*$ never performs worse in accuracy than the simple addition.
Specifically,
\begin{enumerate}
\item if $y=\hat{l}_1+ \hat{l}_2$, these two functions show equivalent accuracy;
\item Otherwise, if $y\neq\hat{l}_1+ \hat{l}_2$, the optimized weighted combination performs better.
\end{enumerate}
\label{prop:specific_1}
\end{proposition}

Proposition \ref{prop:specific_1} indicates the superiority of the weight mechanism over the common additive function in the specific case. The principle behind it is quite simple, as the additive function is a special case of the weighted combination function, i.e., all weight parameters equal $1$. A detailed proof is given in
\ref{appendix:proof_specific_1}. Moreover, the weighted combination function performs better than the additive function unless the additive function already achieves the best. However, the performance of simple addition is usually far away from the optimum in practice.

\begin{proposition}
When $N>2$, there exist infinite groups of the weight parameters optimizing the weighted combination function without residual. With optimized parameters, the optimized combination function performs better in accuracy than the simple addition as long as $y\neq\sum_i\hat{l}_i$.
\label{prop:general_1}
\end{proposition}

As shown in Proposition \ref{prop:general_1}, the superiority of the weight mechanism stays valid for the case with more components. However, the large feasible region of $\{w_i,\forall i\in \{1,\cdots,N\}\}$ reduces the underlying consistency of estimations. We take two measures to reduce the size of the feasible region. First, we incorporate the summing-up constraint $\sum_i w_i=N$ to reduce the dimension of the feasible region by 1. In addition, we restrict the feasible region of each weight parameter in the application, which is proved effective through extensive numerical experiments.

\subsubsection{Impact on Prediction Biases of Components}
\label{sec:weight-2} 

In addition to the overall forecasting accuracy, we are curious about the quality of modified prediction on the components. Specifically, remind that $l_i$ and $\hat{l}_i$ respectively denote the actual value and estimate of component $i$, and $w_i$ is the corresponding weight parameter. Then we want to know that under which condition $|w_i\hat{l}_i-l_i|<|\hat{l}_i-l_i|$, i.e., the modified estimate is less biased than the original prediction . Similar to the analysis in Section \ref{sec:weight-1}, we start with the specific case with $N=2$ for the convenience of mathematics.


\begin{proposition}
When $N=2$, let $-i$ label the other component of component $i$, and
$g(\hat{l}_i)=\hat{l}_i^2+\hat{l}_i\left(y-3\hat{l}_{-i}-2 l_i\right)+2l_i\hat{l}_{-i}.$
Then we have: 
\begin{enumerate}
\item  When $y>\hat{l}_1+\hat{l}_2$, the modified estimate with optimal weight $w_i^*\hat{l}_i$ is less biased than the original prediction $\hat{l}_i$ if and only if $g(\hat{l}_i)<0$;
\item  Otherwise, when $y<\hat{l}_1+\hat{l}_2$, $w_i^*\hat{l}_i$ is less biased than $\hat{l}_i$ if and only if $g(\hat{l}_i)>0$.
\end{enumerate}
\label{prop:specific_2}
\end{proposition}

Proposition \ref{prop:specific_2} shows the conditions of biase reduction for separate components when $N=2$. The variation of biases for component relies on the relative relationship between $y$ and $\hat{l}_1+\hat{l}_2$ (i.e., the prediction with additive combination function), as well as a quadratic function $g(\hat{l}_i)$, which is strictly convex over $\hat{l}_i$.
When the additive combination function under-estimates the actual value, i.e., $y>\hat{l}_1+\hat{l}_2$, the component $i$ is improved after multiplying $w_i^*$ when $g(\hat{l}_i)<0$. Otherwise, when $y<\hat{l}_1+\hat{l}_2$, the additive combination over-estimates the actual value, and the component $i$ is improved with $w_i^*$ if and only if  $g(\hat{l}_i)>0$.

\begin{figure}[htbp]
	\centering
	\caption{The Impact of Weight Mechnanism on Prediction Biases of Components  ($N=2$)}
	\subfloat[$y<\frac{3l_1-l_2}{2}$]{\label{fig:3_larger}\includegraphics[width=6cm]{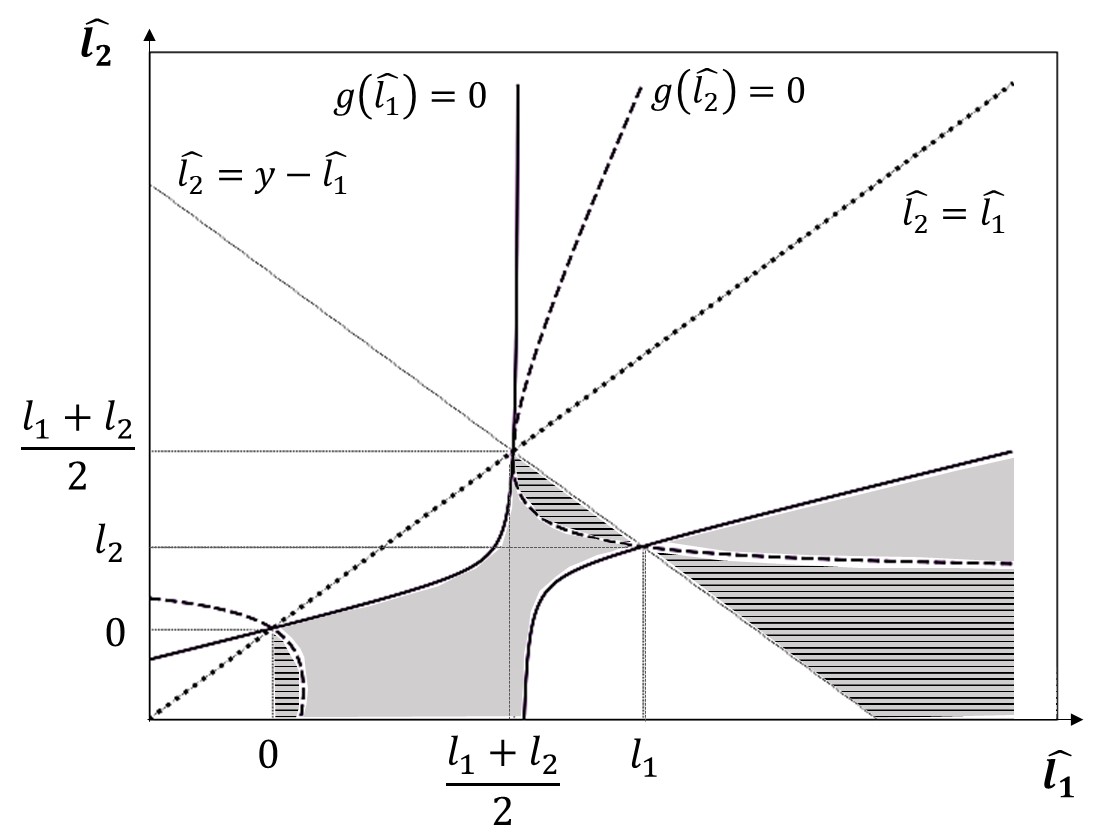}}\quad \quad
\subfloat[$y\geq\frac{3l_1-l_2}{2}$]{\label{fig:3_smaller}\includegraphics[width=6cm]{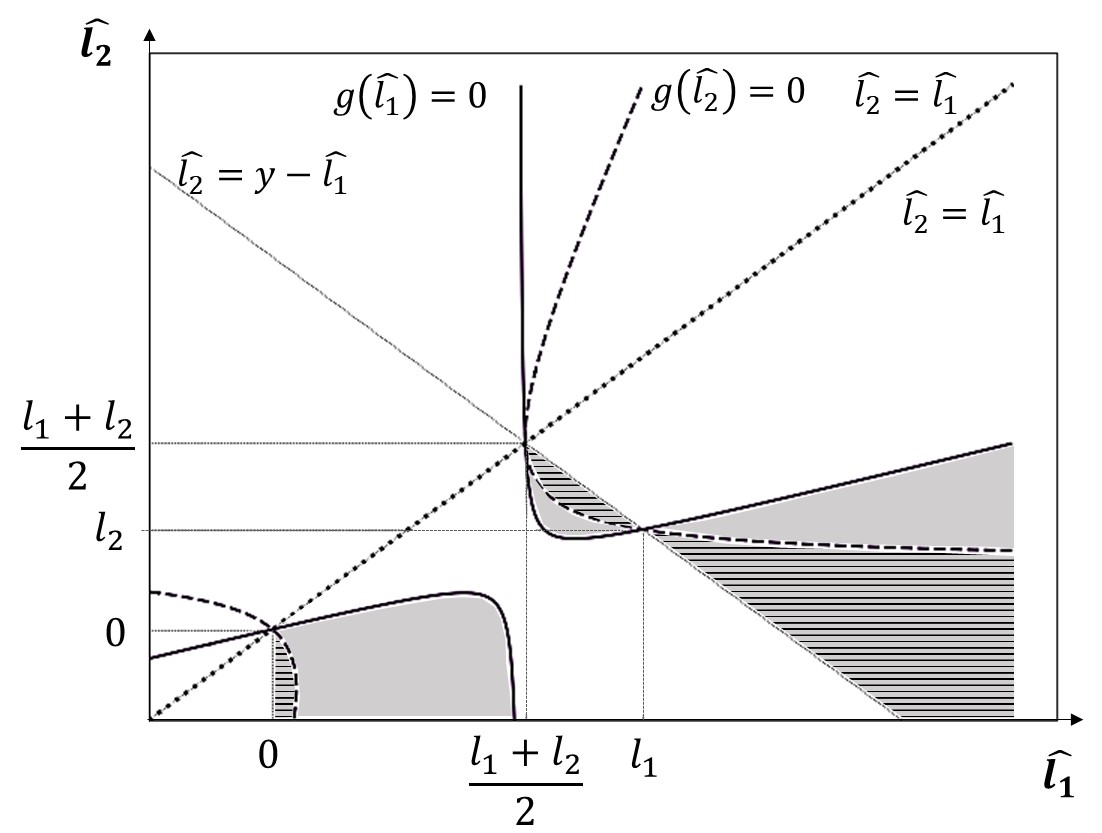}}\\[-0.1cm]
	\includegraphics[width=12.5cm]{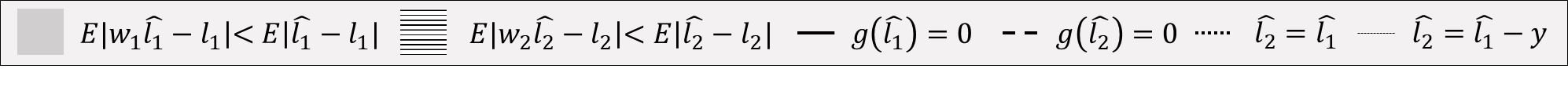}
\label{fig:specific_2}
\vspace{-0.6cm}
\end{figure}

We then explore the conditions for both components to improve with the optimized weights. However, $g(\hat{l}_i)$ and $g(\hat{l}_{-i})$ are two connected quadratic functions, making the joint inequities intricate to solve. Moreover, the mathematical expressions in Proposition \ref{prop:specific_2} are messy, making the solutions for joint improvement difficult to interpret directly.
Therefore, we illustrate the conditions in Figure \ref{fig:specific_2}, where the shaded region denotes the improvement of component $1$, and the area with horizontal lines indicates reduced biases of constituent $2$.
The mathematical conditions for both components improve after the weight mechanism is given by Corollary \ref{coro:specific_2}, illustrated by the shaded area with horizontal lines in Figure \ref{fig:specific_2}. Consistent with Section \ref{sec:weight-1}, we assume $\hat{l}_1>\hat{l}_2$ here.

\begin{corollary}
In the specific case with $N=2$, suppose $\hat{l}_i>0 \ \forall i=\{1,2\}$. Then we have:
\begin{enumerate}
\item  When $\frac{l_1+l_2}{2}<\hat{l}_1<l_1$,  there exists unique $l_{20}^*\in(l_2,\frac{l_1+l_2}{2})$ making $g(\hat{l}_2)=0$, and both of the components improve after multiplying with the optimal weights if and only if $l_{20}^*<\hat{l}_2<\hat{l}_1-y$.
\item When $\hat{l}_1>l_1$, there exists unique $l_{21}^*\in(0,l_2)$ making $g(\hat{l}_2)=0$, and both of the components improve after multiplying with the optimal weights if and only if $\hat{l}_1-y<\hat{l}_2<l_{21}^*$.
\end{enumerate}
\label{coro:specific_2}
\end{corollary}

\begin{observation}
In the specific case with $N=2$, suppose $\hat{l}_i>0 \ \forall i=\{1,2\},\ \hat{l}_1>\hat{l}_2$. Then when $\hat{l}_1>\frac{l_1+l_2}{2}$, $w^*\in(0,2)$.  Specifically, when $\frac{l_1+l_2}{2}<\hat{l}_1<l_1$, $w^*\in(1,2)$; 
when $\hat{l}_1>l_1$, $w^*\in(0,1)$.
\label{coro:specific_3}
\end{observation}

Generally speaking, there are two scenarios that both components improve, (1) when $l_1$ is over-estimated and $l_2$ is under-estimated, and the opposite, (2) when $l_1$ is under-estimated and $l_2$ is over-estimated. In both cases, the original biases for these components are opposite, and the absolute percentages of biases are comparable. Besides, the joint improvement of components is more likely to occur when the additive combination $\hat{l}_1+\hat{l}_2$ is not far from the actual value $y$. This situation is essentially reasonable from the deep logic of the weighted mechanism, i.e., the redistribution function. On the contrary, if both components are over- or under-estimated, it is impossible to reduce the biases simultaneously. Under the optimized scenarios in Corollary \ref{coro:specific_2}, Observation \ref{coro:specific_3} shows the feasible region of  $w^*$ when $N=2$, which is quite intuitive.
When both of the components are positive, $w^*$ should be in $(0,2)$ to ensure $w_i>0,\ \  \forall i\in\{1,2\}$.

\begin{proposition}
For the general case with $\hat{l}_i>0$,  $|w_i\hat{l}_i-l_i|<|\hat{l}_i-l_i|$ if and only if:
\begin{enumerate}
\item $\hat{l}_i<l_i$, and $1<w_i<\frac{2l_i}{\hat{l}_i}-1$;
\item Or $\hat{l}_i>l_i$, and $\frac{2l_i}{\hat{l}_i}-1<w_i<1$.
\end{enumerate}
\label{prop:general_2}
\end{proposition}

\begin{figure}[!htbp]
\vspace{-0.5cm}
		\centering
	\caption{The Impact of Weights on Prediction Biases of Components ($\hat{l}_i>0,\ l_i>0$)}
 \includegraphics[width=2.2in]{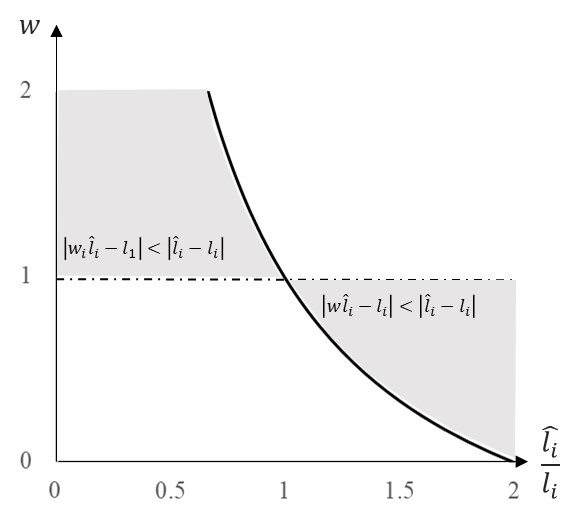}		\label{fig:biase}
\vspace{-0.3cm}
\end{figure}

We finally turn to the general case when $N\geq 2$. Without loss of generality, we focus on the case of $\hat{l}_i>0$ here. Proposition \ref{prop:general_2} presents the general condition for single component in this case, and the graphical illustration is given by Figure \ref{fig:biase}. Specifically, when the component is under-estimated (i.e., $\hat{l}_i<l_i$), the weight mechanism would reduce the forecasting biase when $w_i$ is a bit larger than $1$; and the case of over-estimated (i.e., $\hat{l}_i>l_i$) is just the opposite. The other threshold of $w_i$ besides $1$ is positively related to the relative size of  $l_i/\hat{l}_i$. When $\hat{l}_i<l_i$(or $\hat{l}_i>l_i$), the available region of $w_i$ for reduced biases decreases(increases) with $\hat{l}_i$ when $l_i$ is fixed.
When $\hat{l}_i=0$, the weight mechanism does not work as the multiplication stays 0.
And finally, if $\hat{l}_i<0$ (which is less likely to happen), the results are opposite, and the details are given in Proposition \ref{prop:general_2_negative} in \ref{appendix:proof_general_2}.

\begin{conjecture}
For the general case with $N$ components and non-negative estimates,  it is more likely to happen that  the weight mechanism improves the estimation of all components when
\begin{enumerate}
\item $y$ is relatively close to $\sum_j\hat{l}_j$,
\item and the optimal weights $w^*_i\ \forall i\in\{1,\cdots, N\}$ are in a moderate interval around 1.
\end{enumerate}
\label{conj:general}
\end{conjecture}

 Note that we can only give a general condition with $w_i$ when $N$ is not limited, as the optimal solutions are not unique. However, taking the results from the particular case of $N=2$ and Proposition \ref{prop:general_2}, we have Conjecture \ref{conj:general} for the general conditions of joint components improvement.
It suggests that the estimations on components are more likely to improve by the weight mechanism when the weight parameters are close to $1$. As the exact values of components are unavailable, we validate the conjecture through the overall accuracy in numerical experiments. Specifically, we restrict the feasible region of $w_i$ to $[1-\frac{\alpha}{N},1+\alpha-\frac{\alpha}{N}]$, where $\alpha\in(0,N)$ is a manually determined parameter. Under Conjecture \ref{conj:general}, the overall accuracy would be better when $\alpha$ is moderate, as the estimations on components should be more accurate.

\subsubsection{Discussion}

To summarize, we theoretically analyze the impacts of the weight mechanism without residual on the overall accuracy and components' prediction errors. For the specific case of $N=2$, we derive the unique solution of optimal weights and present the mathematical conditions to reduce the prediction errors of both components. However, for the more general case of $N>2$, the weight parameters minimizing the overall prediction gap are infinite. Inspired by the special case, we guess the component estimates may be less biased when the optimal weights are in a moderate interval around 1. As the exact values of components are unavailable, we believe black-box ML rather than statistical methods would be more appropriate for calculating the weights. To reduce the difficulty of the calculation, we reduce the feasible region of weights by incorporating a summing-up constraint  and forcing each weight within a certain range around 1.

While we only analyze the weighted mechanism without residual here, the main results for the general case still hold for the case with residual. The only difference is replacing $\hat{y}$ and $y$ by $\hat{y}-\varepsilon$ and $y-\epsilon$, where $\varepsilon$ and $\epsilon$ denote the estimate and actual value of the residual, respectively. The incorporation of residual would enlarge the feasible region of alternative parameters and hence increase the calculation complexity. However, it also increases the possibility of bias reduction with improved accuracy, especially when the original prediction with the additive function $y_{additive}$ is less satisfying in accuracy.

\subsection{Concrete Models for Implementation}
\label{subsec:concrete_models}

We finally introduce the implementation details of the W-R algorithm in numerical experiments. Section \ref{subsec:stage_2} presents the MQ-CNN-based module for stage 2, which is committed to improving the forecasting accuracy through the weighted combination function. Sections \ref{subsec:practical} and \ref{subsec:public} describe the specific decomposition algorithms in stage 1. Specifically, we introduce the custom algorithm for the sales decomposition of JD.com for the practical case, and apply the classic STL algorithm to the forecast on the public electricity dataset.

\subsubsection{Stage 2: The ML Module for Improvement}
\label{subsec:stage_2}
\begin{figure}[!htbp]
\vspace{-0.5cm}
		\centering
	\caption{Illustration of the Concrete Model in Stage 2}
 \includegraphics[width=5in]{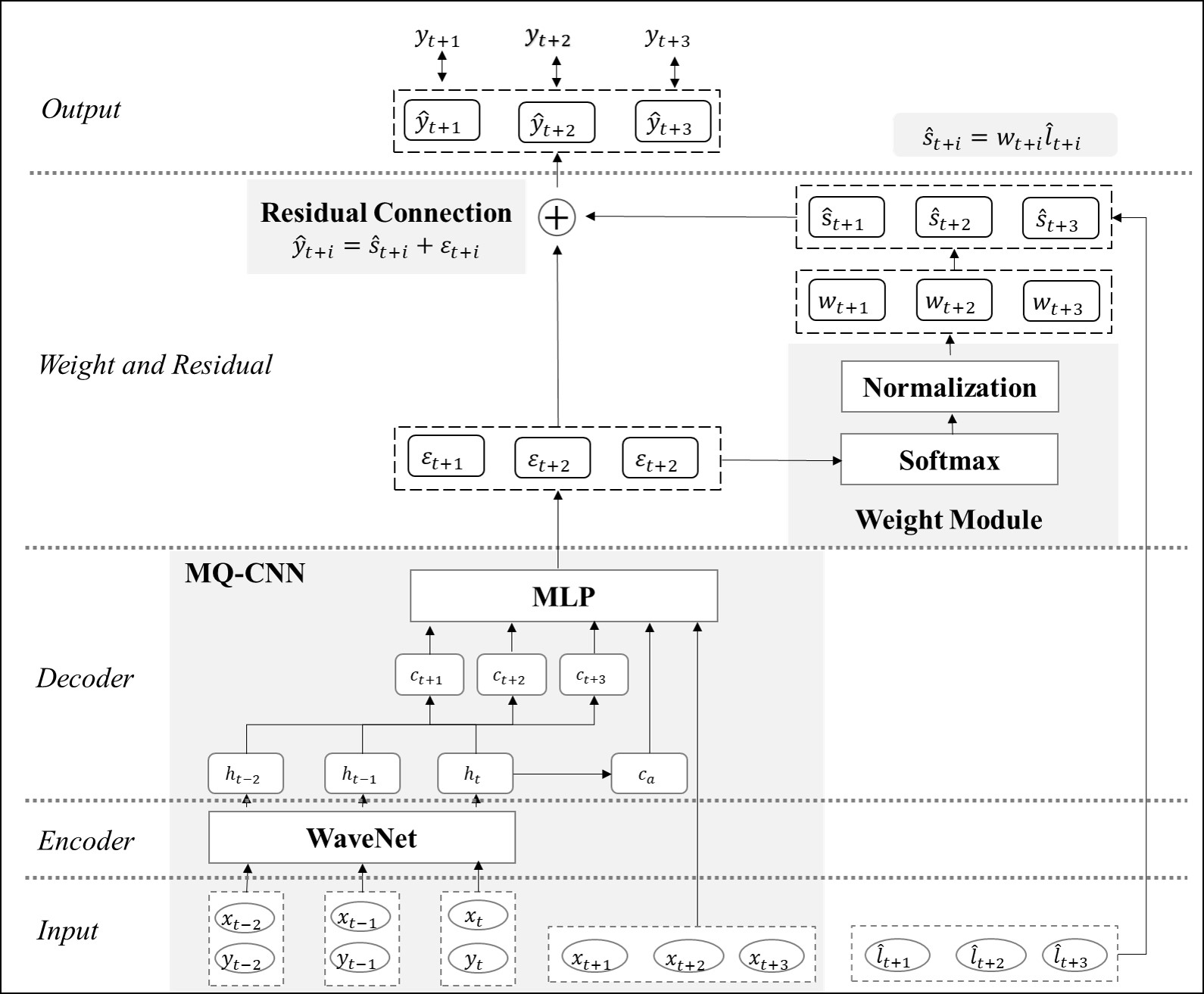}		\label{fig:global}
\vspace{-0.5cm}
\end{figure} 
In stage 2, we utilize the MQ-CNN proposed by \citet{Wen2017} to process empirical data and a weight module with residual connection to generate the weighted combination function for accuracy improvement. Let $T$ denote the length of historical data, then Figure \ref{fig:global} presents the detailed structure when $T=H=3$.

\textbf{\textit{Basic ML Module of MQ-CNN}} The lower-left corner of Figure \ref{fig:global} details the MQ-CNN module.
The historical features $(\bm{x},\bm{y})=\{(x_{t-k},y_{t-k}),\ \forall k \in \{0,\cdots,T-1\}\}$  are first put into a WaveNet module as encoder.  The WaveNet algorithm is a CNN-based auto-regression algorithm widely applied in time series forecasting, and we refer the readers to  \citet{WaveNet2016} if interested. Afterwards, the outputs of WaveNet,  $\bm{h} =\{h_{t-k},\ \forall k \in \{0,\cdots,T-1\}\}$, are sent into  two streams. In the first stream,  the algorithm transforms $h_t$ to $c_a$, which stays stable for different forecasting periods and describes the global information about the historical time series. Meanwhile, the output vector $\bm{h}$  is concatenated and then transformed into vector $\bm{c}$ of size $H$, i.e.,  $\bm{c}=\{c_{t+j},\ \forall j\in\{1,\cdots,H\}\}$. Individual elements within this vector present the historical information for specific forecasting periods and therefore are considered local historical information. For example, $c_{t+j}$ presents the information for horizon $t+j$, and $c_{t+j}$ is likely to differ with $c_{t+k}$ when $k\neq j$. 
And finally, above variables as well as additional features (such as the date of forecast, denoted by $x_{t+j}\ ,\ \forall j\in\{1,\cdots,H\}$) are put into an MLP module to forecast the residuals  $\bm{\varepsilon}=\{\varepsilon_{t+j},\ \ \forall j\in\{1,\cdots,H\}\}$.


\textbf{\textit{Weight Module}}  With the residuals as input, the following module aims to output the weights that apply to the preliminary decompositions. Remind
that we incorporate the summing-up constraint to reduce the dimension of the feasible region for weight parameters. The weight module achieves this through two layers. First, we transform the input residuals through a Softmax layer and get the original weight $w^{(0)}_{i,t+j}$ for component $i$ and horizon $t+j$. Mathematically,
\begin{eqnarray*}
\\[-0.9cm]
w^{(0)}_{i,t+j}=Softmax(\bm{\varepsilon}_{t+j})&=&\frac{exp(\varepsilon_{i,t+j})}{\sum_k exp(\varepsilon_{k,t+j})},
\end{eqnarray*}
where $exp(\cdot)$ denote the exponential function with natural logarithm $e$. Obviously, $\sum_i\hat{w}^{(0)}_{i,t+j} =1$. Then we normalize $w^{(0)}_{i,t+j}$ and get the modified weights by
\begin{eqnarray}
\notag \\[-1cm]
w_{i,t+j}=\alpha w^{(0)}_{i,t+j}+1-\frac{\alpha}{N},
\\[-1cm] \notag
\label{eq:w}
\end{eqnarray}
where $N$ is the number of components from preliminary decomposition, $\alpha$ is a pre-determined parameter controlling the size of the weight interval.
Then under Eq.~(\ref{eq:w}), we have $\sum_iw_{i,t+j}=N, \ \forall j\in\{1,\cdots,H\}$, hence the summing-up constraint is satisfied.
Moreover, Eq.~(\ref{eq:w}) alsp restrict the feasible region of $w_{i,t+j}$  to $[1-\frac{\alpha}{N},1+\alpha-\frac{\alpha}{N}]$, of which the size increases with $\alpha$.
Finally, the modified weights and the residuals, together with the preliminary predictions, combine to output the final overall prediction, i.e.,
\begin{eqnarray*}
\\[-1.1cm]
\hat{y}_{t+j}&=&w_{i,t+j} \hat{l}_{i,t+j} +\varepsilon_{t+j},\ \forall i\in \{1,\cdots,N\},\ j \in \{1,\cdots,H\}.
\\[-0.9cm]
\end{eqnarray*}

\subsubsection{Stage 1: Custom Sales Decomposition in the Practical Case}
\label{subsec:practical}

As noted before, we implement the W-R algorithm in two cases.
The practical case is the sales prediction for JD.com, one of China's largest online retailers.
Following \citet{CHAN4CAST2005,Abolghasemi2020} and practical experience, we decompose the sales into baseline, promotion- and festival-related components in the first stage, which follow different logic and are generated individually.

\textbf{\textit{Baseline}}  The baseline indicates regular sales without promotion or specific events. We utilize the FFORMA algorithm proposed by \citet{FFORMA2018}, which use XGBoost from \citet{Chen2016} to generate a series of weight parameters on alternative forecasting methods.
 Let $w_{m}(f_n)$ and $L_{m}(f_n)$ denote the weight and loss for the $n$th sample with features $f_n$ and forecasting model $m$, then the XGBoost model   outputs $w_{m}(f_n)$ to minimize $\textit{Loss}=\sum_{n} \sum_{m} w_{m}(f_n)\times L_{m}(f_n)$ in training. When predicting with input feature $f_k$, the weight parameter $w_m(f_k)$ is the automatically generated by the XGBoost model, and the final prediction for baseline sales, $\hat{l}_b(f_k)$, is a weighted combination of $\hat{l}_{m}(f_k)$, the original forecasting results of alternative models. Mathematically,
\begin{eqnarray}
\notag \\[-0.7cm]
\hat{l}_b(f_k)&=&\sum_{m}  w_m(f_k) \times \hat{l}_{m}(f_k).
\label{eq:FFORMA_prediction}
\\[-0.7cm] \notag
\end{eqnarray}

As we only focus on regular sales, we omit the spikes induced by promotions or events within the sales data in the training of the numerical experiments. The alternative forecasting models include the moving average (MA), weighted MA, ETS, and ARIMA methods, all of which are classical statistical algorithms with wide application in practice \citep{Chun2021}.
More details on the alternative statistical methods and FFORMA algorithm are given in \ref{appendix:base0} and \ref{appendix:base}, and we refer the readers to \citet{Hyndman2021} and \citet{FFORMA2018} for more details if interested.

\begin{figure}[!htbp]
\vspace{-0.2cm}
		\centering
	\caption{Casual Relationship among Price, Promotion Type and Sales}

 \includegraphics[width=3in]{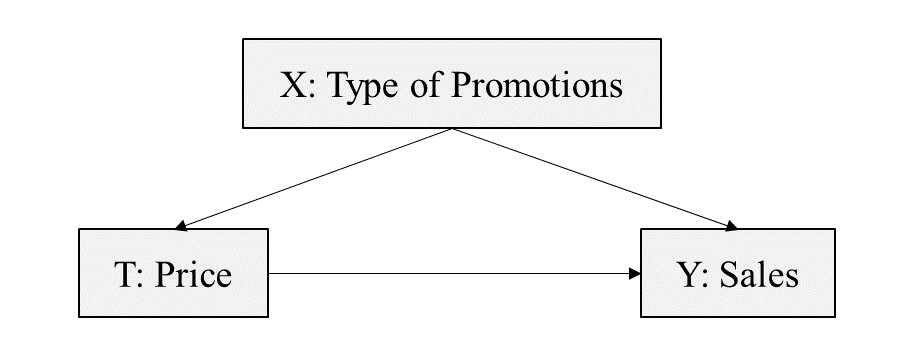}		\label{fig:promotion}
\vspace{-0.5cm}
\end{figure}

\textbf{\textit{Promotion}} Promotion is an intricate marketing activity with the highest expenditure in the retail industry \citep{vanHeerde2008}. To capture the complex impacts of price discount and promotion type on sales (as shown in Figure \ref{fig:promotion}), we use Double Machine Learning (DML for short) to eliminate the impacts of promotion type as well as other confounding variables in predicting the promotion-induced sales. Similarly, we briefly introduce the model here and refer the readers to \citet{DML2017} for more detail.

The theoretical foundation of DML is simple mathematics in causal inference. Suppose $X$ is the control variable (i.e., promotion type), $Y$ is the outcome variable, $T$ is the treatment variable, $W$ denotes other confounding variables (e.g., date, original price, and weekdays in our experiment). Then following the partially linear regression model in \citet{Robinson1988},  we have
\begin{eqnarray*}
\\[-0.9cm]
Y&=&\theta(X,W) T+f_0(X,W)+\varepsilon_0, \\ T&=&f_1(X,W)+\varepsilon_1,
\\[-0.9cm]
\end{eqnarray*}
where $\varepsilon_0$, $\varepsilon_1$ are unobserved noises such that  $E[\varepsilon_0|X,W,T]=0$, and  $E[\varepsilon_1|X,W,\varepsilon_0]=0$. The goal is to estimate the conditional average treatment effect of sample $x$, i.e., $\hat{\theta}(x)=E[\theta(X,W)|X=x]$. After some simple math calculations, \citet{Oprescu2018} found that
\begin{eqnarray*}
\\[-0.9cm]
E[\tilde{Y}|X,\tilde{T}]&=&\theta(X) \cdot\tilde{T},
\\[-0.9cm]
\end{eqnarray*}
where $\tilde{Y}=Y-f_0(X,W)$ and $\tilde{T}=T-f_1(X,W)$. Therefore  
 we can get $\hat{\theta}(X)$ by regressing $\tilde{Y}$ on $\tilde{T}$ locally around $X=x$. Then the detailed procedures for DML calculation are as follows.
\begin{enumerate}
\item Use ML algorithm to estimate $Y$ and $T$ based on $X$ and $W$, and get $\hat{f}_0(X,W)$  and $\hat{f}_1(X,W)$;
\item Calculate the residuals $\tilde{Y}$ and $\tilde{T}$;
\item Use another ML algorithm to regress $\tilde{Y}$ on $\tilde{T}$ and get the estimation $\hat{\theta}(X)$.
\end{enumerate}

Following the LinearDML in \citet{econml}, we use gradient-boosting decision tree (GBDT) from 
\citet{Friedman2001} in Stage 1 and linear regression  in Stage 3 in practice.  Moreover, we apply $Y=log(sales)$ and $T=log(price)$ in the experiment, which makes $\hat{\theta}(X)$ exactly the price elasticity over sales \citep{Kaplan2011}.



\textbf{\textit{Festival}} The final module captures the impacts of shopping festivals. These festivals are pre-determined in levels by the marketing department of JD.com, e.g., \textit{S level} for the \textit{618 and 1111 Shopping Festivals}, \textit{A level} for the \textit{Valentine's Day and School Season}.
The festival-related sales $l_{f}$ are forecasted by a simple linear regression model, and the prediction $\hat{l}_{f}=\beta \hat{l}_b$, 
where $\beta$ is the festival factor, and $\hat{l}_b$ is the prediction of baseline sales. According to the historical data during previous festivals of the same level, $\beta=\frac{l_{actual}-\hat{l}_b}{\hat{l}_b}$, where $l_{actual}$ is the actual sales, and $\hat{l}_b$ is the corresponding prediction on baseline.

\subsubsection{Stage 1: STL Algorithm for the Public Case}
\label{subsec:public}

To validate the effectiveness and scalability of the W-R algorithm, we apply the classical STL algorithm \citep[Seasonal-Trend decomposition procedure based on Loess, proposed by][]{Cleveland1990} on a public electricity dataset. In accordance with previous sections, we briefly introduce the Loess method, a basic method within STL algorithm, and the general procedure of the STL algorithm here, and we refer the readers to \ref{appendix:stl} and \citet{Cleveland1990} for more details.

\textbf{\textit{The Loess Method}} \  Loess algorithm is a weighted variant of the K-Nearest Neighbor (KNN) method with additional parameters $q$, $d$, and $\delta$. Mathematically, suppose it aims to predict $y(x)$ with samples $\{(x_i,y_i),\forall i\in\{1,\cdots,T\}\}$. Then following the KNN algorithm, we choose the $q$ samples closest to $x$, denoted by $S$. For each $x_i\in S$, we calculate the weights $\bm{v}$ through a given function $W(\cdot)$ and serves as a normalization parameter. Mathematically,
\begin{eqnarray*}
\\[-0.7cm]
v_i(x)&=&\delta W\left(\frac{|x_i-x|}{\lambda_q(x)}\right), \ \textit{where}\  \lambda_q(x)=\left\{ \begin{array}{ll}max(|x_i-x|),\ \forall x_i\in S \ &\textit{ if } q\leq T \\[0.1cm] max(|x_i-x|)\frac{q}{n},\ \forall x_i\in S\ &\textit{ if } q>T\end{array} \right.,
\\[-0.4cm]
\end{eqnarray*}
where  $W(\cdot)$ is often modeled as the tricube weight function (i.e., $W(x)=(1-x^3)^3$).
And finally, we fit $\bm{v}$ by a polynomial function of degree $d$, and get the result of Loess algorithm.

\textbf{\textit{Calculation Procedure of STL}} \ 
In STL algorithm, $\bm{l}=\bm{l_T}+\bm{l_S}+\bm{l_R}$, where $\bm{l}$ is the time series, $\bm{l_T}$, $\bm{l_S}$ and $\bm{l_S}$ denote the trend, seasonality and residual components, respectively. Specifically, the STL algorithm mainly includes two iterative loops, an outer loop to calculate $\bm{l_R}$, and an inner loop to estimate $\bm{l_T}$ and $\bm{l_S}$. The general calculation procedure is as follows.
\begin{enumerate}
\item Initialization: set $\bm{l_T}=0$.
\item Inner Loop:
\begin{enumerate}
\item Use Loess to regress $\bm{l}-\bm{l_T}$, and get $\bm{C}$;
\item Use MA and Loess to filter $\bm{C}$ and get $\bm{L}$, and update $\bm{l_S}=\bm{C}-\bm{L}$;
\item Use Loess to regress $\bm{l}-\bm{l_S}$ and get $\bm{l_T}$.
\end{enumerate}
\item Outer Loop: Calculate $\bm{l_R}=\bm{l}-\bm{l_T}-\bm{l_S}$.
\item Repeat the inner and outer loops until the termination condition.
\end{enumerate}

\section{Numerical Experiment and Results}
\label{sec:result}
This section presents the numerical results. Section \ref{subsec:design} introduce the data and experiment settings, and Section \ref{sebsec:results} reports the numerical results. Finally, in Section \ref{subsec:post-hoc-1}, we analyze the underlying mechanism from the algorithm's outputs, providing evidence for the algorithm's effectiveness.

\subsection{Experiment Design}
\label{subsec:design}
\subsubsection{Loss Function and Evaluation Indicators}
As mentioned in Section \ref{subsec:structure}, the W-R algorithm predicts quantiles in multiple forecasting horizons. Accordingly, following \citet{Wen2017}, we adopt the quantile loss function
\begin{eqnarray*}
Loss (y,\hat{y})=p(y-\hat{y} )_+ +(1-p) (\hat{y}-y)_+,
\end{eqnarray*}
where $(\cdot)_+=max(0,\cdot)$, $y$ is the actual value, $\hat{y}$ is the forecasted value of quantile $p$. In the experiments, we select $p=0.5$, the most widely applied parameter, and the minimizer of loss is the median of the predictive distribution \citep{5}. Remember that $t$ denotes the forecasting time, $H$ is the length of the forecasting horizon, then we evaluate the models' performance through the two indicators, Root Median Square Error (RMSE for short) and P50\underline{ }QL. Mathematically,
\begin{eqnarray*}
\\[-0.5cm]
\textit{RMSE}&=&\sqrt{\textit{average}\left(\sum_{i,t} \left(y_{i,t}-\hat{y}_{i,t}\right)^2\right)}, \ 
\textit{P50\underline{ }QL}=\frac{\sum_{i,t}|\sum_{j=1}^H \hat{y}_{i,t+j}-\sum_{j=1}^H {y}_{i,t+j}|}{2\sum_{i,t}|\sum_{j=1}^H \hat{y}_{i,t+j}|},
\\[-0.5cm]
\end{eqnarray*}
where  $y_{i,t+j}$ and $\hat{y}_{i,t+j}$ denote the actual and forecasted value for sample $i$ in period $t+j$, respectively. These indicators are essentially similar to the \textit{Root Mean Square Error} and \textit{Weighted Mean Absolute Percentage Error}, just replacing the mean with the median. Specifically, RMSE is a 
general  indicator evaluating the detailed performance on separate periods, and P50\underline{ }QL is an aggregated indicator widely applied in multi-horizon forecasting \citep[e.g., in][]{Wen2017, DeepAR2020}. 

\subsubsection{Data and Descriptive Statistics}
\label{subsec:data}

\begin{figure}[htbp]
\vspace{-0.2cm}
	\centering
	\caption{Distribution of the Time Series in Numerical Experiments}
\vspace{-0.1cm}
	\subfloat[Distribution of JD.com's Sales (Practical Data)]{\label{fig:salse_history}\includegraphics[width=7.5cm]{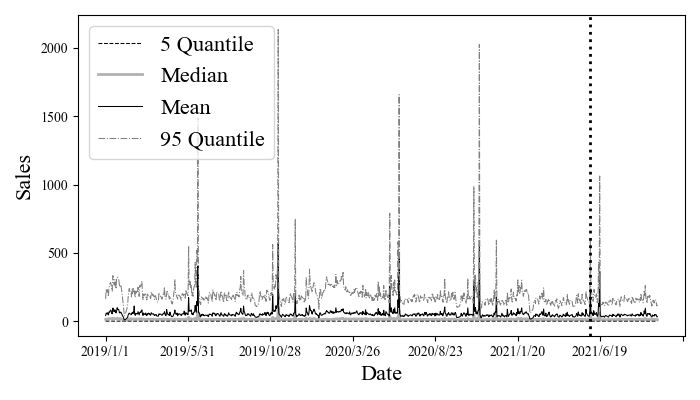}}\quad
	\subfloat[Distribution of Electricity Loads  (Public Data)]{\label{fig:elec_history_0}\includegraphics[width=7.5cm]{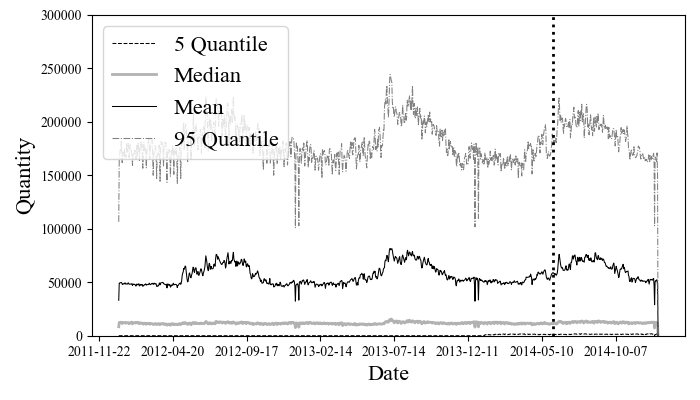}}
\vspace{-1cm}
\end{figure}

In the practical case, we utilize the data of 16,973 individual products from 298 categories in a given distribution center of JD.com, and the specific product information is anonymous for security reasons. Figure \ref{fig:salse_history} presents the distribution of individual sales over time, where the dotted vertical line divides the training and test sets.
As the products are launched for sale at different times, we randomly select $72$ samples for each product  to avoid the influence of bias from data in training.
Generally speaking, the sales are significantly right-skewed, and the products with higher sales are more volatile. Specifically, the indicator of 95 quantiles has two significant peaks every year, corresponding to the \textit{618} and \textit{11-11 Shopping Festivals} that occur in June and November. Therefore, the custom decomposition algorithm of JD.com is more effective than other general algorithms, as it considers the significant influencing factors of sales, i.e., the promotion and festival.

The public dataset, meanwhile, is the electricity loads data provided by \citet{25}. The dataset consists of the time series data of 370 electrical clients from early 2011 to late 2014 and has been widely applied in relevant literature \citep[e.g.,][]{ERTUGRUL2016,Shih2019,CHEN2020,DeepAR2020}. We calculate the daily time series as the forecast target in our experiment. As shown in Figure \ref{fig:elec_history_0}, the electricity loads are also right-skewed and highly volatile, similar to the sales in Figure \ref{fig:salse_history}. However, there are no significant peaks induced by specific events. Hence we utilize the STL method as the preliminary decomposition algorithm. As both the STL and W-R algorithms consider the residual component, here we only use the trend and seasonality components from STL as the initial predictions on components.

\subsubsection{Features and Parameters}
As listed in Table \ref{table:practical_features}, Section \ref{subsec:features}, the features for the practical case include historical data on sales, prices, and promotions, and extra information on time and products. Specifically, in the practical case, we consider four actual promotion types in JD.com: direct discount, seckilling (a time/quantity-limited discount), $X$ off on purchases over $Y$ for a variety of products, and bundled discount, in addition to the classical historical and additional features. As for the experiment on public datasets, we only utilize the original time series, client information, and time features as other data are unavailable.

The parameters, as listed in Table \ref{table:parameters}, \ref{subsec:features}, are all common parameters in similar researches.
Briefly speaking, for the practical case, we use the historical information of the past $72$ days to forecast the sales of the following $24$ days from $1$st, June to September 2021. On the public dataset, we predict the time series in a month starting from three pre-determined forecasting dates, $1$st June, $1$st July, and $1$st August 2014, with the historical data before the date as the training set. To ensure the robustness of all results, we replicate each experiment $10$ times and report the average of evaluation indicators. In each experiment, we train the model from $10$ to $50$ times and report the evaluation indicators with the best performance. We utilize tensorflow-based python programming in both experiments, and the open-source packages used are listed in Section \ref{appendix:packages}.

\subsection{Numerical Results}
\label{sebsec:results}
\subsubsection{The Impact of $\alpha$}

Before evaluating the accuracy, we need to determine $\alpha$ in Eq. (\ref{eq:w}), the parameter controlling the scope of weights. Inspired by the theoretical results, we restrict $\alpha\in[0,N]$, and the results on practical and public data are given by Tables \ref{table:alpha} and \ref{table:alpha_pub}, respectively.

\begin{figure}[!htbp]
\begin{minipage}{.48\linewidth}
\captionof{table}{Model Performance with Different $\alpha$ \\[-0.2cm] (Practical Data, $N=3$)}
\label{table:alpha}
\small
{\def\arraystretch{1}
\begin{tabular}{cccc}
\hline
\hline
 \textbf{Value of $\bm{\alpha}$}& \textbf{RMSE} & \textbf{P50\underline{ }QL} & \textbf{Scope of Weight} \\
\hline
$0$& $2.1659$&$0.2089$& 1 \\
$0.5$& $2.1544$&$0.2067$& $[0.83,1.33]$\\
$\bm{1}$& $\bm{2.0396}$&$\bm{0.2001}$&$\bm{[0.67,1.67]}$\\
$2$& $2.1595$&$0.2062$&$[0.33,2.33]$\\
$3$& $2.1583$&$0.2092$&$[0,3]$\\
\hline
\hline
\end{tabular}
}
\end{minipage}
\hfill
\begin{minipage}{.48\linewidth}
\captionof{table}{Model Performance with Different $\alpha$ \\[-0.2cm] (Public Data, $N=2$)}
\label{table:alpha_pub}
\small
{\def\arraystretch{1}
\begin{tabular}{cccc}
\hline
\hline
 \textbf{Value of $\bm{\alpha}$}& \textbf{RMSE} & \textbf{P50\underline{ }QL} & \textbf{Scope of Weight} \\
\hline
$0$& $0.7106$&$0.0464$& 1 \\
$0.5$& $0.6821$&$0.0303$& $[0.75,1.25]$\\
$\bm{1}$& $\bm{0.4682}$&$\bm{0.0261}$&$\bm{[0.50,1.50]}$\\
$1.5$& $1.5263$&$0.0651$&$[0.25,1.75]$\\
$2$& $1.4264$&$0.0853$&$[0,2]$\\
\hline
\hline
\end{tabular}
}
\end{minipage}
\vspace{-0.3cm}
\end{figure}

Note that the weights are equal to $1$ when $\alpha=0$, and the model is just a simple addition with an additive residual. The feasible region of weights increases with $\alpha$, and the W-R algorithm achieves the best performance in both cases when $\alpha=1$. This phenomenon validates the Conjecture \ref{conj:general} in Section \ref{sec:weight}. First, $\alpha$ should be positive to ensure the effectiveness of the weight mechanism. In addition, when $\alpha$ makes the weights in a proper, moderate interval around $1$, the estimates on components are less biased, making the total prediction more accurate. Moreover, by comparing the results from practical and public cases, the improvements led by the W-R algorithm and the weight mechanism are both more significant when the simple additive function shows poor performance.

\subsubsection{Model Performance}
\label{subsec:preformance}

\begin{figure}[!htbp]
\vspace{-0.1cm}
\begin{minipage}{.49\linewidth}
\vspace{-0.65cm}
\captionof{table}{Comparison on Model Performance \\[-0.1cm]  (Practical Data)}
\label{table:results}
\quad
\small
{\def\arraystretch{0.8}
\begin{tabular}{lcc}
\hline
\hline
\textbf{Algorithm}                     &\textbf{RMSE}        & \textbf{P50\underline{ }QL}       \\
\hline
MQ-CNN                         & 2.5107 & 0.2324 \\
\makecell[l]{Preliminary Decomposition \uppercase\expandafter{\romannumeral1}\\[-0.2cm] \textit{(Additive Combination)}}    & 2.4104 & 0.2193  \\
\makecell[l]{Preliminary Decomposition \uppercase\expandafter{\romannumeral2}\\[-0.2cm] \textit{(Simple MLP Combination)}}    &2.3303    & 0.2115   \\
W-R Algorithm ($\alpha=1$)          & 2.0396 & 0.2001 \\
DeepAR                        & 2.6334    & 0.2610   \\
N-BEATS                        & 2.7522 & 0.3017       \\
\hline
\hline  
\end{tabular}
}
\end{minipage}
\hfill \quad
\begin{minipage}{0.49\linewidth}
\caption{Model Performance Over Time \\[-0.1cm]  (Practical Data)}
\label{fig:practical}
\vspace{-0.1cm}
 \centerline{\includegraphics[width=7.5cm]{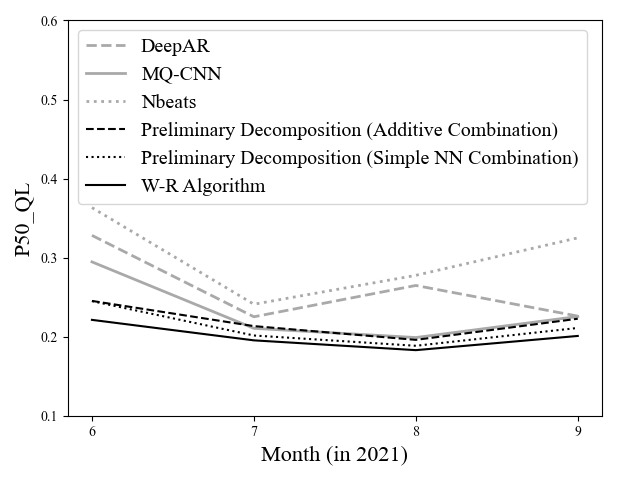}}
\end{minipage}
\vspace{-0.8cm}
\end{figure}

Table \ref{table:results} and Figure \ref{fig:practical} present the performance comparison among models in the practical case. In addition to the W-R algorithm with $\alpha=1$, we also consider three streams of benchmarks here. The first stream serves as a general benchmark and includes two popular existing algorithms, DeepAR and N-BEATS.
The second stream aims to validate the idea of combining ML and preliminary decompositions and consists of the W-R algorithm's basic modules, including the MQ-CNN and the simple addition of the custom sales decomposition algorithm, named \textit{Preliminary Decomposition \uppercase\expandafter{\romannumeral1}}. The final benchmark is \textit{Preliminary Decomposition \uppercase\expandafter{\romannumeral2}}, a variation of the W-R algorithm which uses a simple MLP to separately output the weights and residual, to validate the effect of residual connection.

The results are satisfying in every sense. First, we validate the importance of custom decomposition for sales. Even the simplest version, \textit{Preliminary Decomposition \uppercase\expandafter{\romannumeral1}} with simple additive combination, achieves better performance than pure ML algorithms, especially during big promotions (i.e., the \textit{618 Shopping Festival} in June). On the basis of such a good prediction, the W-R algorithm steadily improves the forecasting accuracy, ranging from $2.60\%$ to $84.79\%$ in absolute P50\underline{ }QL, and $0.08$ to $0.62$ in absolute RMSE by month (see Table \ref{table:practice_detail} for specific statistics). Overall, the relative accuracy improvement is $8.76\%$ in P50\underline{ }QL, and $15.38\%$ in RMSE. While the approach using simple MLP to output the weights and residuals also improves the preliminary decomposition, the extent of improvement is less significant than our algorithm. Therefore, the specific structure of residual connection in the W-R algorithm is necessary, in addition to the idea of modification through weight and residual.

\begin{figure}[h]
\begin{minipage}{.48\linewidth}
\vspace{-0.6cm}
\centering
\captionof{table}{Comparison on Model Performance \\[-0.1cm]  (Public Data)}
\label{table:results2}
\vspace{0.2cm}
\small
{\def\arraystretch{1}
\begin{tabular}{lcc}
\hline
\hline
\textbf{Algorithm}                     &\textbf{RMSE}        & \textbf{P50\underline{ }QL}       \\
\hline
STL with  Additive Combination           & 2.2967 & 0.1186  \\
W-R Algorithm  ($\alpha=1$)      &0.4682&0.0261    \\
MQ-CNN&0.6265&0.0364\\
DeepAR&1.2097&0.0710\\
N-BEATS          & 1.1203   & 0.0698      \\
\hline
\hline  
\end{tabular}
}
\end{minipage}
\hfill \quad
\begin{minipage}{0.48\linewidth}
\caption{Model Performance Over Time \\[-0.1cm] (Public Data)}
\label{fig:public}
\vspace{-0.1cm}
 \centerline{\includegraphics[width=6.5cm]{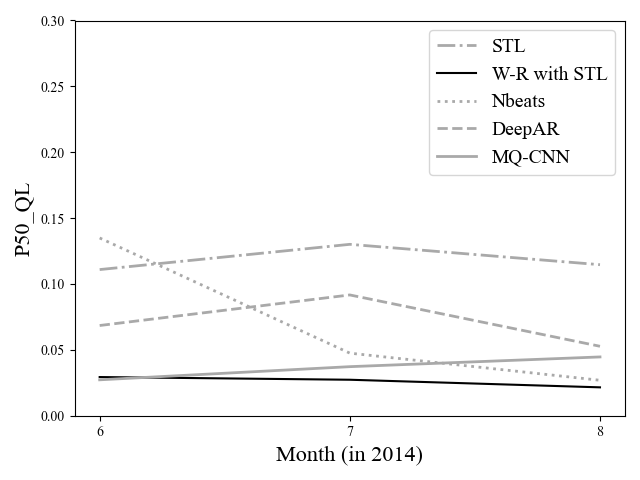}}
\end{minipage}
\vspace{-0.8cm}
\end{figure}

We further check the robustness of the W-R algorithm's superiority through experiments on the public dataset. As shown in Table \ref{table:results2} and Figure \ref{fig:public}, the numerical results on public datasets validate the poor performance of STL, which is not surprising, and the significant improvement induced by the W-R algorithm, which is of $9.25\%$($77.99\%$) in absolute(relative) P50\underline{ }QL and $1.8285$($79.62\%$) in absolute(relative) RMSE. The improvement in accuracy certainly comes from the ML structure, i.e., the MQ-CNN algorithm, which performs the second best among the alternatives.

\subsection{Post-hoc Analysis on Undelying Mechanism}
\label{subsec:post-hoc-1}

\begin{figure}[!htbp]
  \centering
  \caption{Distribution of the Weights and Residuals (Practical Data)}
\vspace{-0.1cm}
	\label{fig:data_distribution_practical}
    \subfloat[Weights on Baseline]{\includegraphics[width=0.25\textwidth]{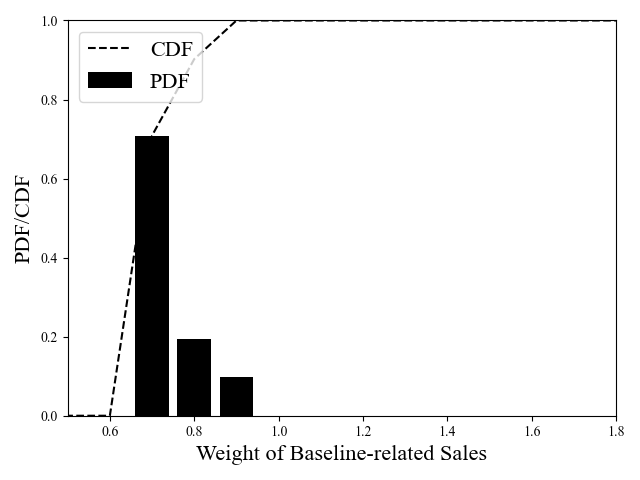}} 
	\subfloat[Weights on Promotion ]{\includegraphics[width=0.25\textwidth]{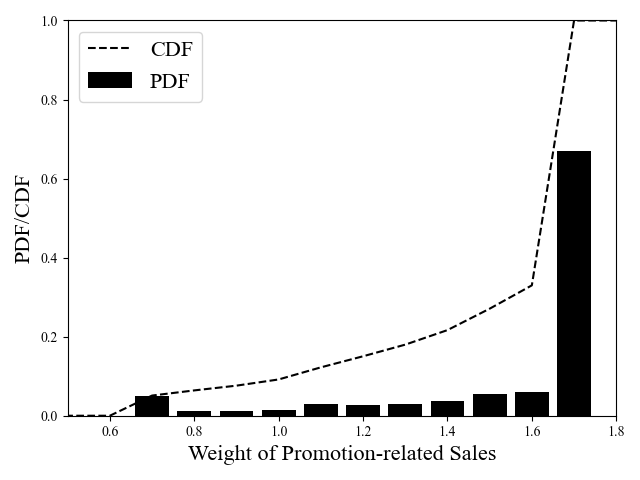}} 
    \subfloat[Weights on Festival]{\includegraphics[width=0.25\textwidth]{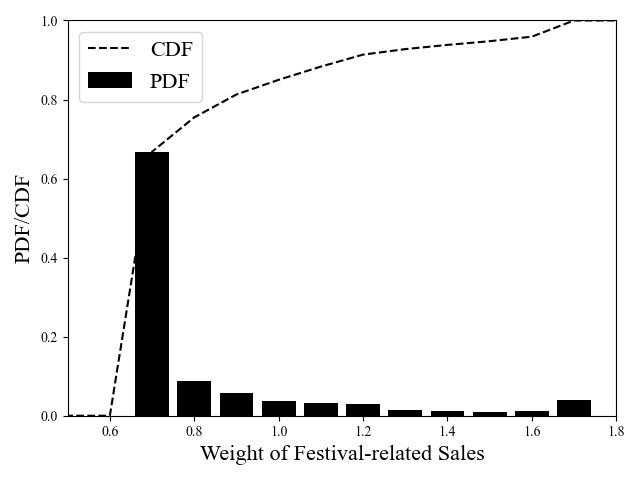}}
\subfloat[Residuals]{\includegraphics[width=0.25\textwidth]{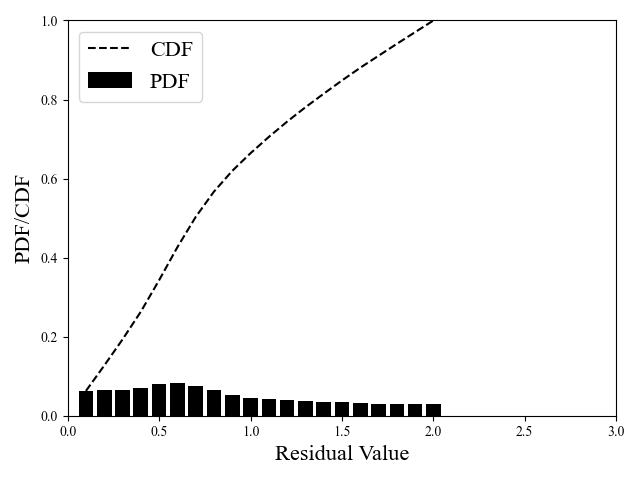}} 
\vspace{-0.8cm}
\end{figure}

We finally check the underlying mechanism of the W-R algorithm in the numerical experiments through post-hoc analysis. We start with the practical sales forecast case and draw the weight parameters' probability distribution in Figure \ref{fig:data_distribution_practical}. Note that the weights of baseline sales are all less than $1$; that is, the W-R algorithm thinks the custom decomposition algorithm overestimates the baseline sales. The idea is essentially appropriate. Remind that we predict the baseline through a combination of naive statistical methods. While we have deleted the spikes closely related to exogenous variables, it is likely that we missed some small promotions, making the original baseline include some sales from small promotions.

The weights on the promotions and festivals are almost opposite in distribution. In general, the W-R algorithm amplifies the effects of promotions on sales but reduces those of festivals. This phenomenon makes sense as the simple linear regression model for the festival does not exclude the impacts of simultaneous promotions. As we all know, it isn't easy to distinguish between highly relevant promotions and festivals; however, the W-R algorithm has a different view of these two components compared to human predictions. Finally, following a nearly uniform distribution from $0$ to $2$, the residuals are relatively stable compared to the weights and relatively small in contrast to other components. Therefore, we infer that the improvement of forecasting accuracy mainly comes from the redistribution of components through the weight parameter in the practical case.

\begin{figure}[!htbp]
  \centering
  \caption{Distribution of the Weights and Residuals (Public Data)}
\vspace{-0.1cm}
	\label{fig:data_distribution_public}
	    \subfloat[Weights on Trend]{\includegraphics[width=0.25\textwidth]{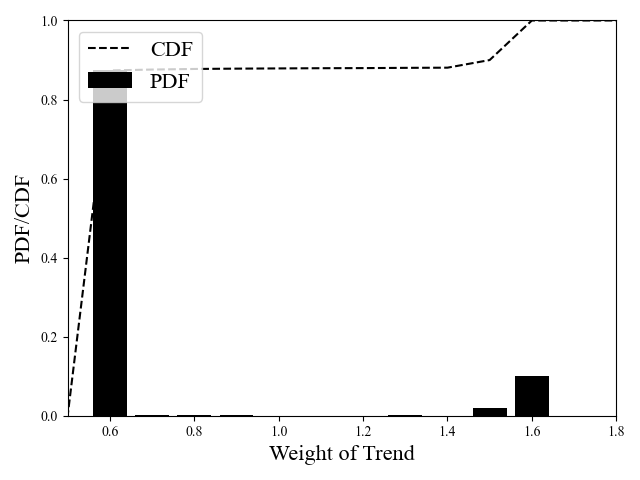}\label{fig:trend_pub}} 
	\subfloat[Weights on Seasonality]{\includegraphics[width=0.25\textwidth]{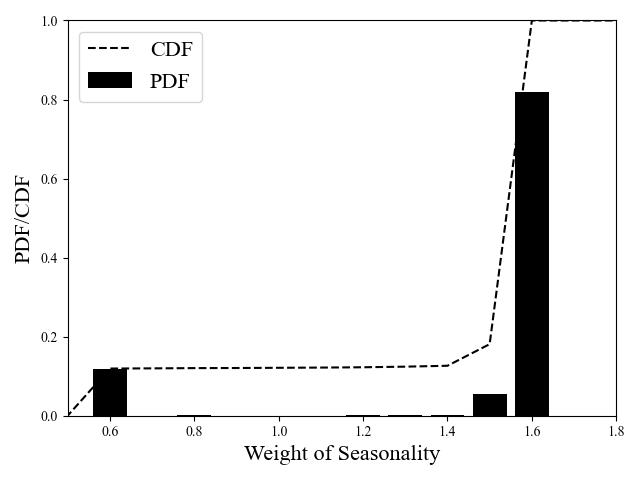}\label{fig:seasonality_pub}} 
	\subfloat[Residuals]{\includegraphics[width=0.25\textwidth]{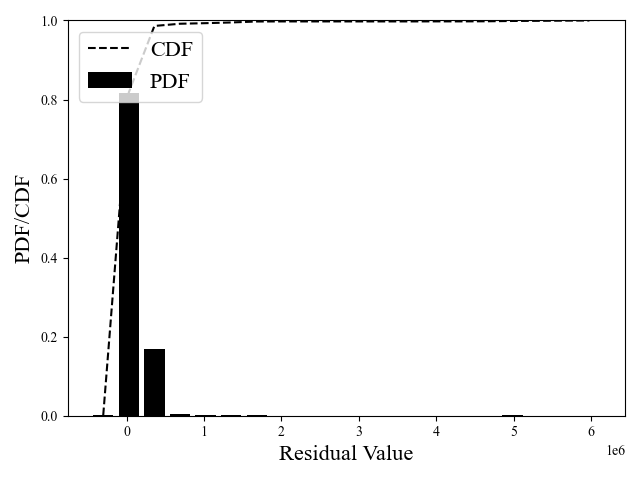}\label{fig:residual_pub}}
\vspace{-0.8cm}
\end{figure}

For the experiment on the public dataset, we depict the distribution of output parameters in Figure \ref{fig:data_distribution_public}. First, as illustrated in Figures \ref{fig:trend_pub} and \ref{fig:seasonality_pub}, the W-R algorithm modifies the weights of trend and seasonality components. In most cases, the trend component is reduced while the seasonality component is magnified. However, the trend and seasonality components do not explain the whole time series. As shown in Figure \ref{fig:residual_pub}, the residuals are large in values, with an average of $39,290.41$. Apparently, we should search for additional explanatory components to improve the existing algorithm, e.g., the unit price for electricity.

\section{Conclusion and Discussion}
\label{sec:conclusion}

This research is inspired by the persisting conflict between accuracy and interpretability in time series forecasts. In response, we first quantitatively define interpretability in data-driven forecasting. The innovative definition includes the interpretability of inputs and the function, making distinct algorithms comparable in interpretability. We then build a unifying framework and systematically review the existing algorithms. We find that while there are several hybrid interpretable ML algorithms, these algorithms usually rely on strong statistical assumptions or theoretical structures, which restricts the potential of ML. In particular, the additive combination function in most decomposition-based algorithms implies an assumption of independence, and the algorithms' performance would be compromised when the components are correlated.

Accordingly, we propose a novel W-R algorithm for time series forecasting, which has specific advantages in multi-dimensional accuracy and interpretability. First, integrating the decomposition-based and ML approaches allows the algorithm to be accurate in extensive cases, even when the preliminary decompositions are inappropriate. Moreover, the theoretical analysis of the weighted mechanism reveals the possibility of reducing the prediction errors of components, implying that the modified estimates on the components from the W-R algorithm are likely to improve with proper restrictions. Finally, in contrast with other hybrid algorithms, the W-R algorithm utilizes ML to generate the parameters in a weighted combination function rather than the direct predictions. According to the innovative definition of interpretability, the weighted combination function is just as interpretable as the simple addition, and hence our algorithm is just as interpretable as other algorithms with additive combination functions.

Compared with existing literature, the W-R algorithm provides an innovative perspective for combining the classic statistical and black-box ML algorithms. Based on a weighted combination function, the algorithm improves the accuracy of decomposition-based algorithms and the interpretability of ML algorithms.
Moreover, the W-R algorithm possesses great generalization ability in the application, and the resulting accurate predictions on components have significant managerial implications for firms. At this point, the W-R algorithm has been incorporated by JD's intelligent forecasting system, and the resulting predictions on promotion and festivals serve as important guides for marketing activities.



\bibliographystyle{informs2014} 
\setlength{\parskip}{0.3cm}
 \let\oldbibliography\thebibliography
 \renewcommand{\thebibliography}[1]{%
 	\oldbibliography{#1}%
 	\baselineskip15pt 
 	\setlength{\itemsep}{4.5pt}
 }
\bibliography{Main_file_2} 




\newpage
\ECSwitch 

\vspace{-0.6cm}
\ECHead{E-Companions}

\vspace{-0.2cm}
\section{Mathematical Proofs for Section \ref{sec:weight}}
\label{appendix:proof}
\subsection{Proof of Proposition \ref{prop:specific_1}}
\label{appendix:proof_specific_1}

 When $N=2$,  the weighted combination function is $\hat{y}= w \hat{l}_1+(2-w)\hat{l}_2$, and the additive combination function is a special case with $w=1$. Then the optimization problem is 
\begin{eqnarray*}
\min_w E[y-\hat{y}]\  \Leftrightarrow \ \min_w L=\left(y-\hat{y}\right)^2=\left(y-w \hat{l}_1-(2-w)\hat{l}_2\right)^2.
\end{eqnarray*}
We then calculate the first and second order of derivatives, i.e.,
\begin{eqnarray*}
\frac{\partial L}{\partial w} &=& 2\left(y-w \hat{l}_1-(2-w)\hat{l}_2\right)\left(-\hat{l}_1+\hat{l}_2\right),\\
\frac{\partial^2 L}{\partial w^2} &=& 2\left(-\hat{l}_1+\hat{l}_2\right)^2\geq 0.
\end{eqnarray*}
 Then when $\hat{l}_1=\hat{l}_2$, $\frac{\partial L}{\partial w}=\frac{\partial^2 L}{\partial w^2}=0$, the objective function stays constant for all $w\in[0,2]$. Otherwise, when $\hat{l}_1\neq\hat{l}_2$,  $\frac{\partial^2 L}{\partial w^2}>0$, the objective function is strictly convex over $w$. 
Let $\frac{\partial L}{\partial w}=0$, we have 
\begin{eqnarray}
w^*=\frac{{y}-2{\hat{l}}_2}{{\hat{l}}_1-{\hat{l}}_2}. \label{eq:optimal}
\end{eqnarray}
 When the simple addition achieves the best performance, $w^*=1$, $y={\hat{l}}_1+{\hat{l}}_2$. Otherwise, $w^*\neq1$, the weighted combination function performs better than the simple addition. 
 $\square$\\

\vspace{-0.5cm}
\subsection{Proof of Proposition \ref{prop:general_1}}
\label{appendix:proof_general_1}

When $N>2$,   the weighted combination function is ${\hat{y}}= w_1 {\hat{l}}_1+{\hat{l}}_2+\cdots+w_{N-1}{\hat{l}}_{N-1}+(N-w_1-w_2-\cdots-w_{N-1}){\hat{l}}_N$, and the additive combination function is a special case with $\bm{w}=\mathbf{1}$. Then the optimization problem is 
\begin{eqnarray*}
\min_{\bm{w}} \left|\bm{{y}-{y}}\right| \  \Leftrightarrow \ \min_{\bm{w}} L&=&\left({y}-{\hat{y}}\right)^2\\
&=&\left(y-w_1 \hat{l}_{1}-w_2\hat{l}_{2}-\cdots-w_{N-1}\hat{l}_{N-1}-(N-w_1-w_2-\cdots-w_{N-1})\hat{l}_{N}\right)^2.
\end{eqnarray*}
 Then obviously, the objective function has a minimum of $0$ when  $y=w_1^* \hat{l}_{1}+w_2^*\hat{l}_{2}+\cdots-w_{N-1}^*\hat{l}_{N-1}+(N-w_1^*-w_2^*-\cdots-w_{N-1}^*)\hat{l}_{N}$. The optimal weights $\bm{w^*}\in \mathbb{R}^{N-2}$ if there's not any other additional constraint.  $\square$\\
\textbf{Example: The specific case of numerical experiment } In the numerical experiment of our research, we further restrict $w_i^*\in[1-\frac{\alpha}{N},1-\frac{\alpha}{N}+\alpha]\ \forall i$, where $\alpha\in[0,N]$. Then in the practical case of sales prediction,  $N=3$,  the feasible region of weights is $\{w_1^*,w_2^*,w_3^*\}\in \mathbb{R}$, which satisfying
\begin{eqnarray*}
w_1^*\in[1-\frac{\alpha}{N},1-\frac{\alpha}{N}+\alpha],\ 
w_2^*=\frac{y+w_1^*(\hat{l}_{3}-\hat{l}_{1})-3\hat{l}_{3}}{\hat{l}_{2}-\hat{l}_{3}},\ 
w_3^*=N-w_1^*-w_2^*.
\end{eqnarray*}

\subsection{Proof of Proposition \ref{prop:specific_2}}
\label{appendix:proof_specific_2}

\begin{lemma}
If $\hat{l}_i(w_i-1)\left((w_i+1)\hat{l}_i-2l_i\right)<0$, $|w_i\hat{l}_i-l_i|<|\hat{l}_i-l_i|$, the modified estimation $w_i\hat{l}_i$ is less biased than the original prediction $\hat{l}_i$.
\label{lemma:general_2}
\end{lemma}
\textbf{Proof:}
\begin{eqnarray*}
|w_i\hat{l}_i-l_i|<|\hat{l}_i-l_i| \ & \Leftrightarrow & \ (w_i\hat{l}_i-l_i)^2-(\hat{l}_i-l_i)^2<0\\ 
& \Leftrightarrow & (w_i\hat{l}_i-\hat{l}_i)(w_i\hat{l}_i+\hat{l}_i-2l_i) <0\\
& \Leftrightarrow & \hat{l}_i(w_i-1)((w_i+1)\hat{l}_i-2l_i) <0.\ \square
\end{eqnarray*}

When $N=2$, the weight parameter for component $1$ $w^*_1=w^*=\frac{{y}-2{\hat{l}}_2}{{\hat{l}}_1-{\hat{l}}_2}$. According to Lemma \ref{lemma:general_2}, when $\hat{l}_i  >0$, $w_1^*\hat{l}_1$ is less biased than $\hat{l}_1$ if and only if
\begin{eqnarray*}
(w_1^*-1)((w_1^*+1)\hat{l}_1-2l_1) <0 &\Leftrightarrow &\frac{\hat{l}_1}{(\hat{l}_1-\hat{l}_2)^2}\left(y-\hat{l}_1-\hat{l}_2\right) \left(\hat{l}_1^2+\hat{l}_1 (y-3\hat{l}_2-2l_1)+2 l_1 \hat{l}_2\right)<0\\
&\Leftrightarrow & \left(y-\hat{l}_1-\hat{l}_2\right) g(\hat{l}_1)<0.
\end{eqnarray*}
Similarly, the weight parameter for component $2$ $w^*_2=2-w^*=\frac{2{\hat{l}}_1-{y}}{{\hat{l}}_1-{\hat{l}}_2}$, and 
\begin{eqnarray*}
(w_2^*-1)((w_2^*+1)\hat{l}_2-2l_2) <0 &\Leftrightarrow &\frac{\hat{l}_2}{(\hat{l}_1-\hat{l}_2)^2}\left(y-\hat{l}_1-\hat{l}_2\right) \left(\hat{l}_2^2+\hat{l}_2 (y-3\hat{l}_1-2l_2)+2 l_2 \hat{l}_1\right)<0\\
&\Leftrightarrow & \left(y-\hat{l}_1-\hat{l}_2\right) g(\hat{l}_2)<0.
\end{eqnarray*}
Taking together, we have Proposition \ref{prop:specific_2}. Specifically, when $y>\hat{l}_1+\hat{l}_2$(or $y<\hat{l}_1+\hat{l}_2$), component $i$ is less biased after multiplying with the $w_i^*$ if and only if $g(\hat{l}_i)<0$(or $g(\hat{l}_i)>0$). $\square$

\subsection{Proof of Proposition \ref{prop:general_2}}
\label{appendix:proof_general_2}

Following Lemma \ref{lemma:general_2}, when $\hat{l}_i>0$, $|w_i\hat{l}_i-l_i|<|\hat{l}_i-l_i| \Leftrightarrow  (w_i-1)((w_i+1)\hat{l}_i-2l_i) <0$. Mathematically, let $T(w_i)=(w_i-1)((w_i+1)\hat{l}_i-2l_i)=\hat{l}_i w_i^2-2l_i w_i -\hat{l}_i+2l_i$, which is quadratic over $w_i$. Moreover, as $(-2l_i)^2-4\hat{l}_i (-\hat{l}_i+2l_i)=4(\hat{l}_i-l_i)^2\geq 0$, $T(w_i)=0$ has two distinct solutions as long as $\hat{l}_i\neq l_i$. We denote these two solutions by $w_i^{[*1]}$ and $w_i^{[*2]}$ with $w_i^{[*1]}<w_i^{[*2]}$.
\begin{enumerate}
\item When $l_i>\hat{l}_i$, $w_i^{[*1]}=1$ and $w_i^{[*2]}=\frac{2l_i}{\hat{l}_i}-1$. The first scenario in Proposition \ref{prop:general_2} is proved. 
\item Otherwise, when $l_i<\hat{l}_i$, $w_i^{[*1]}=\frac{2l_i}{\hat{l}_i}-1$ and $w_i^{[*2]}=1$. The second scenario in Proposition \ref{prop:general_2} is proved. $\square$
\end{enumerate}

Here we also briefly discuss the case when $\hat{l}_i<0$. In this case, $|w_i\hat{l}_i-l_i|<|\hat{l}_i-l_i| \Leftrightarrow (w_i-1)((w_i+1)\hat{l}_i-2l_i) >0 \Leftrightarrow T(w_i)>0$, and the two solutions of $T(w_i)=0$ stay the same. We then have Proposition \ref{prop:general_2_negative} as follows.
\begin{proposition}
When $\hat{l}_i<0$, $|w_i\hat{l}_i-l_i|<|\hat{l}_i-l_i|$ if and only if:
\begin{enumerate}
\item $l_i>\hat{l}_i$, and $w_i<1$ or $w_i>\frac{2l_i}{\hat{l}_i}-1$;
\item Otherwise, $l_i<\hat{l}_i$, and $w_i<\frac{2l_i}{\hat{l}_i}-1$ or $w_i>1$.
\end{enumerate}
\label{prop:general_2_negative}
\end{proposition}
It is worth mentioning that this scenario rarely happens in practice for the following two reasons. First, we can always modify the negative estimates to the positive ones by adding a positive constant. Secondly, some time series are impossible to be negative due to the physical characteristics. For example, our numerical experiments are built on sales and electricity loads, which are naturally non-negative.

\subsection{Proof of Corollary \ref{coro:specific_2}}
\label{appendix:proof_corollary_specific_2}
\begin{lemma}
In the specific case with $N=2$, suppose $\hat{l}_1>\hat{l}_2$ . Then component $1$ improves as long as component $2$ improves.
\label{lemma:coro_2}
\end{lemma}
\textbf{Proof:}
In the specific case without residual, $y=l_1+l_2$. Following the definition of $g(\hat{l}_i)$, we have
\begin{eqnarray*}
g(\hat{l}_1)-g(\hat{l}_2)&=&(\hat{l}_1-\hat{l}_2)\left(y+\hat{l}_1+\hat{l}_2-2(l_1+l_2)\right)\\
&=&(\hat{l}_1-\hat{l}_2)\left(\hat{l}_1+\hat{l}_2-y\right).
\end{eqnarray*}
Then when $y>\hat{l}_1+\hat{l}_2$, $g(\hat{l}_1)-g(\hat{l}_2)<0$, $g(\hat{l}_1)<0$ as long as $g(\hat{l}_2)<0$. Similarly, when $y<\hat{l}_1+\hat{l}_2$, $g(\hat{l}_1)-g(\hat{l}_2)>0$, $g(\hat{l}_1)>0$ as long as $g(\hat{l}_2)>0$. Taking Proposition \ref{prop:specific_2} together, Lemma \ref{lemma:coro_2} is proved. $\square$

Then according to Proposition \ref{prop:specific_2}, the conditions for both components to improve become:
\begin{enumerate}
\item  $y>\hat{l}_1+\hat{l}_2$ and $g(\hat{l}_2)<0$;
\item Or the opposite, $y<\hat{l}_1+\hat{l}_2$ and $g(\hat{l}_2)>0$.
\end{enumerate}
We analyze the previous two scenarios respectively.\\
\textbf{Scenario 1:} The first scenario requires  $y>\hat{l}_1+\hat{l}_2$ and $g(\hat{l}_2)<0$, where
\begin{eqnarray*}
g(\hat{l}_2)<0  \ & \Leftrightarrow & \ y<3\hat{l}_1+2l_2-\hat{l}_2-2\frac{l_2 \hat{l}_1}{\hat{l}_2}.
\end{eqnarray*}
Therefore 
\begin{eqnarray*}
  \hat{l}_1+\hat{l}_2<y<3\hat{l}_1+2l_2-\hat{l}_2-2\frac{l_2 \hat{l}_1}{\hat{l}_2}\  \to \ \hat{l}_1+\hat{l}_2<3\hat{l}_1+2l_2-\hat{l}_2-2\frac{l_2 \hat{l}_1}{\hat{l}_2}  \ \to  \ \hat{l}_2>l_2,
\end{eqnarray*}
and scenario 1 has no-empty solution only if $\hat{l}_2>l_2$.
And we have the following restrictions on the feasible region of $\hat{l}_1$ and $\hat{l}_2$,
\begin{eqnarray*}
y=l_1+l_2>\hat{l}_1+\hat{l}_2, \  \hat{l}_2>l_2 \ & \Leftrightarrow & \  \hat{l}_1<l_1+l_2-\hat{l}_2<l_1,\\
y=l_1+l_2>\hat{l}_1+\hat{l}_2,\  \hat{l}_1>\hat{l}_2   \ & \Leftrightarrow & \  \hat{l}_2 <\frac{l_1+l_2}{2}.
\end{eqnarray*}
Moreover, when $\hat{l}_1 >\frac{l_1+l_2}{2}$, $\hat{l}_1>y-\hat{l}_1$, and the feasible region of $\hat{l}_2$ is $(l_2,y-\hat{l}_1)$. 
As $g(\hat{l}_2)$ is strictly convex over $\hat{l}_2$, and 
\begin{eqnarray*}
g(\hat{l}_2=l_2) &=&l_2 (l_1-\hat{l}_1)>0,\\
g(\hat{l}_2=y-\hat{l}_1) &=& 2(l_1 -\hat{l}_1)(l_1+l_2-2\hat{l}_1)<0 \textit{ as } \hat{l}_1 >\frac{l_1+l_2}{2} \textit{ and } l_1>\hat{l}_1 .
\end{eqnarray*}
Therefore, there exists unique solution $l_{20}^*\in(l_2, \frac{l_1+l_2}{2})$ making $g(l_{20}^*)=0$. When $y>\hat{l}_1+\hat{l}_2$, $g(\hat{l}_2)<0$ is  equivalent to  $l_{20}^*<\hat{l}_2<y-\hat{l}_1$. This scenario corresponds to the first case in Corollary \ref{coro:specific_2}. \\
\textbf{Scenario 2:} The second scenario requires $y<\hat{l}_1+\hat{l}_2$ and $g(\hat{l}_2)>0$, where
\begin{eqnarray*}
g(\hat{l}_2)>0  \ & \Leftrightarrow & \ y>3\hat{l}_1+2l_2-\hat{l}_2-2\frac{l_2 \hat{l}_1}{\hat{l}_2}.
\end{eqnarray*}
Therefore 
\begin{eqnarray*}
  \hat{l}_1+\hat{l}_2>y>3\hat{l}_1+2l_2-\hat{l}_2-2\frac{l_2 \hat{l}_1}{\hat{l}_2}\  \to \ \hat{l}_1+\hat{l}_2>3\hat{l}_1+2l_2-\hat{l}_2-2\frac{l_2 \hat{l}_1}{\hat{l}_2}  \ \to  \ \hat{l}_2<l_2,
\end{eqnarray*}
and scenario 2 has no-empty solution only if $\hat{l}_2<l_2$.
And we have following restrictions on the feasible region of $\hat{l}_1$ and $\hat{l}_2$,
\begin{eqnarray*}
y=l_1+l_2<\hat{l}_1+\hat{l}_2, \ \hat{l}_2<l_2 \ & \Leftrightarrow & \ \hat{l}_1>l_1+l_2-\hat{l}_2>l_1.
\end{eqnarray*}
As $g(\hat{l}_2)$ is strictly convex over $\hat{l}_2\in(0,l_2)$, and
\begin{eqnarray*}
g(\hat{l}_2=0) &=&2 l_2 \hat{l}_1>0\\
g(\hat{l}_2=l_2) &=&l_2 (l_1-\hat{l}_1)<0,
\end{eqnarray*}
we have the second case in Corollary \ref{coro:specific_2}. That is, there exists unique solution $l_{21}^*\in(0,l_2)$ making $g(l_{21}^*)=0$. When $y<\hat{l}_1+\hat{l}_2$, $g(\hat{l}_2)<0$ is equivalent to $y-\hat{l}_1<\hat{l}_2<l_{21}^*$. \\
\textbf{Additional Discussion for Scenario 1:}
Finally, while not mentioned in Corollary \ref{coro:specific_2}, we also briefly discuss the situation when $\hat{l}_1 \leq\frac{l_1+l_2}{2}$ in Scenario 1. In this case, the feasible region of $\hat{l}_2$ becomes $(l_2,\hat{l}_1)$, with $g(\hat{l}_2=\hat{l}_1)>0$. Then the $g(\hat{l}_2)<0$ has feasible solutions if and only if
\begin{eqnarray*}
\left\{
\begin{array}{l}
l_2< \frac{3\hat{l}_1+l_2-l_1}{2} \\[0.2cm]
\hat{l}_1> \frac{3\hat{l}_1+l_2-l_1}{2}\\ [0.2cm]
g(\hat{l}_2=\frac{3\hat{l}_1+l_2-l_1}{2} ) <0
\end{array}
\right.
\ \Leftrightarrow \
\left\{
\begin{array}{l}
\hat{l}_1> \frac{l_1+l_2}{3} \\[0.2cm]
\hat{l}_1< l_1-l_2\\ [0.2cm]
9\hat{l}_1^2-2\hat{l}_1(3l_1+l_2)+(l_2-l_1)^2 >0
\end{array}
\right.
\ \Leftrightarrow \
\left\{
\begin{array}{l}
l_1\geq 3l_2 \\[0.2cm]
\hat{l}_1^*<\hat{l}_1<\frac{l_1+l_2}{2}
\end{array}
\right. ,
\end{eqnarray*}
where $\hat{l}_1^*\in (\frac{l_1+l_2}{3},\frac{l_1+l_2}{2})$ is the unique solution making
$9\hat{l}_1^2-2\hat{l}_1(3l_1+l_2)+(l_2-l_1)^2 =0$. As you expect, these inequalities mean strict restrictions. Under these conditions, there exists two distinct solutions $l_{22}^*$ and $l_{23}^*$ satisfying $l_2<l_{22}^*<l_{23}^*<\frac{l_1+l_2}{2}$, $g(\hat{l}_2=l_{22}^*)=0$ and $g(\hat{l}_2=l_{23}^*)=0$. And both of the components improve with the weighted mechanism if and only if $l_{22}^*<\hat{l}_2<l_{23}^*$.

Taking together, when $\hat{l}_1 \leq\frac{l_1+l_2}{2}$, the conditions for both the components improve are
\begin{eqnarray*}
\left\{  
             \begin{array}{l}  
             l_1\geq 3l_2  \\[0.2cm]  
             \frac{l_1+l_2}{3} <\hat{l}_1^*<\hat{l}_1<\frac{l_1+l_2}{2}\\[0.2cm] 
l_2<l_{22}^*<\hat{l}_2<l_{23}^*<\frac{l_1+l_2}{2}
             \end{array}  
\right.  ,
\end{eqnarray*}
which are obviously quite strict. Therefore, we omit this situation  and only focus on the cases with $\hat{l}_1 >\frac{l_1+l_2}{2}$ in the main text. $\square$

\subsection{Detailed Explanation of Observation \ref{coro:specific_3}}
\label{appendix:proof_corollary_specific_3}

\begin{figure}[htbp]
	\centering
	\caption{Contours of $w^*$ when $N=2$}
	\subfloat[$y<\frac{3l_1-l_2}{2}$]{\label{fig:3_larger}\includegraphics[width=6cm]{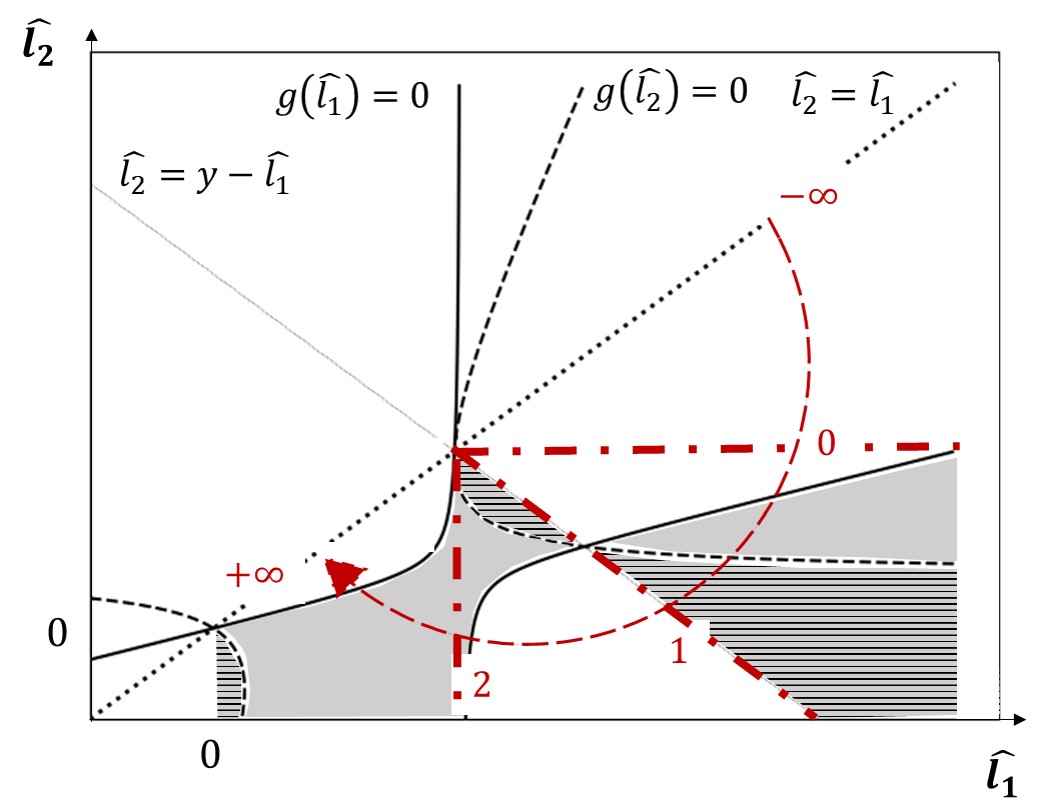}}\quad \quad
\subfloat[$y\geq\frac{3l_1-l_2}{2}$]{\label{fig:3_smaller}\includegraphics[width=6cm]{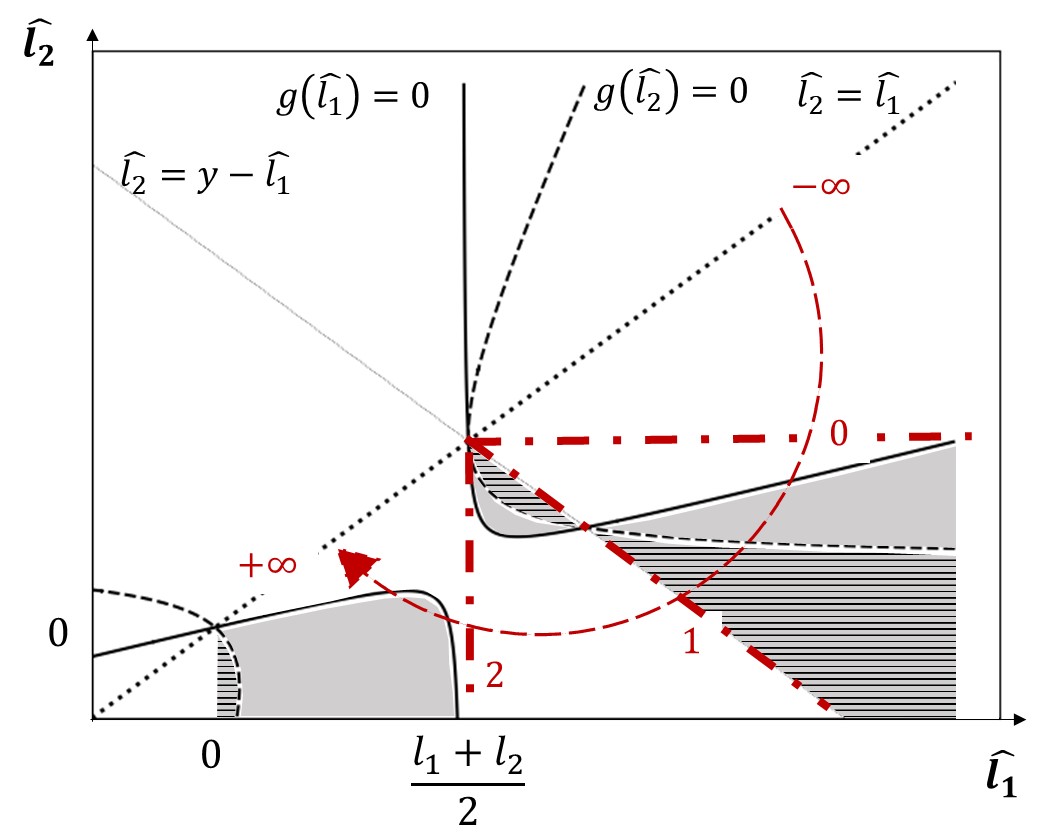}}\\[0.1cm]
	\includegraphics[width=12.5cm]{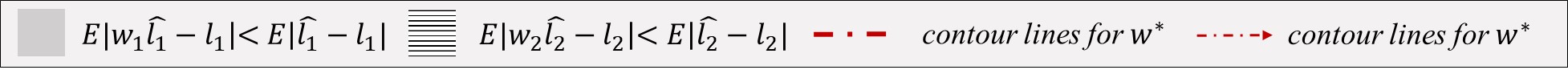}
\label{fig:specific_3}
\end{figure}

Following the definition of $w^*$ when $N=2$ (see Proposition \ref{prop:specific_1}), \begin{eqnarray*}
w^*=\frac{y-2\hat{l}_2}{\hat{l}_1 -\hat{l}_2} \  \Leftrightarrow  \  \hat{l}_2 =\frac{w^* \hat{l}_1-y}{w^* -2} \textit{\ \  if \ \ } w^*\neq2.
\end{eqnarray*}
Mathematically, with given $w^*$,  $\hat{l}_2$ is a linear function over $\hat{l}_1$,  making  the contours for $w^*$ in Figure \ref{fig:specific_2} are straight lines with a fixed point $(\frac{l_1+l_2}{2},\frac{l_1+l_2}{2})$.  Specifically, when $w^*=0$, $1$ and $2$, the corresponding functions are $\hat{l}_2 =\frac{l_1+l_2}{2}$, $\hat{l}_2 =-\hat{l}_1+l_1+l_2$, and $\hat{l}_1 =\frac{l_1+l_2}{2}$, respectively. We draw the distribution of $w^*$ in Figure \ref{fig:specific_3}, as a result, Observation \ref{coro:specific_3} is quite intuitive.

\newpage
\section{Supplement Descriptions on Algorithms in Section \ref{subsec:concrete_models}}
\vspace{0.2cm}
\subsection{Alternative Statistical Algorithms in Section \ref{subsec:practical}}
\label{appendix:base0}
\subsubsection{MA and Weighted MA}
Suppose we want to forecast $y_t$ at time $t$, and the time series data we use are $\bm{y}=\{y_{t-T},\cdots,y_{t-1}\}$. Then  mathematically, 
\begin{eqnarray*}
\textit{For MA: }\hat{y}_t&=&\frac{\sum_{j=1}^{T} y_{t-j}}{T},\\
\textit{For Weighted MA: }\hat{y}_t&=&\frac{\sum_{j=1}^{T} \left(\frac{2(T-j+1)}{T(T-1)}y_{t-j}\right) }{T}=\frac{2 \sum_{j=1}^{T} ((T-j+1)y_{t-j})}{T^2(T-1)}.
\end{eqnarray*}

\subsubsection{Holt-Winters Additive Method (ETS)}
The holt-winters method decomposes the time series $\bm{y}$ into the average (or baseline), trend, and seasonality   components, denoted by $\bm{l}$, $\bm{b}$ and $\bm{s}$, respectively. Suppose $m$ is the frequency of seasonality, $\alpha$, $\beta$ and $\gamma$ are pre-determined parameters between $[0,1]$. Then in the additive ETS model, the prediction on time $t+h$ at time $t$ is 
\begin{eqnarray*}
\hat{y}_{t+h|t}=l_t+h b_t +s_{t+h-m \lfloor\frac{h-1}{m}+1 \rfloor}
\end{eqnarray*}
where $l_t =\alpha(y_t-s_{t-m})  +(1-\alpha)(l_{t-1}+b_{t-1})$, $ b_t=\beta (l_t-l_{t-1})+(1-\beta) b_{t-1}$, and $s_t=\gamma(y_t-l_{t-1}-b_{t-1})+(1-\gamma)s_{t-m}$.

\subsubsection{ARIMA Without and with Seasonality}
ARIMA is based on ARMA, an algorithm combing auto-regression and MA methods. Let $p$ denote the order of the auto-regressive part, $q$ denote the order of the moving average part, then the prediction $\hat{y}_t$ of model \textit{ARMA(p,q)} follows
\begin{eqnarray*}
\hat{y}_t=\mu + \sum_{i=1}^{p} \phi_i \hat{y}_{t-i}+ \sum_{j=1}^{q} \theta_j \varepsilon_{t-j}+ \varepsilon_t,
\end{eqnarray*}
where $\mu$, $\phi_i$ and $\theta_j$ are the parameters to estimate according to the maximum likelihood principle, $\forall i\in\{1,\cdots,p\}, j\in\{1,\cdots,q\}$. When $\bm{y}=\{y_t\}$ is not stationary, the data need to be transformed by differencing $d$ times, and the differenced time series $\hat{y}^{'}_t$ follows a similar form, i.e.,
\begin{eqnarray*}
\hat{y}^{'}_t=\mu + \sum_{i=1}^{p} \phi_i \hat{y}^{'}_{t-i}+ \sum_{j=1}^{q} \theta_j \varepsilon_{t-j}+ \varepsilon_t,
\end{eqnarray*}
and is denoted by model \textit{ARIMA(p,d,q)} as $\hat{y}_t$ is calculated by integration. If the time series also have significant seasonality, we need to make another differencing operation with the order $D$ and lag (or seasonality frequency) $m$. The resulting seasonal ARIMA is denoted as \textit{ARIMA}$(p,d,q)(P,D,Q)_m$, where $P$, $D$, and $Q$ are the parameters of the seasonal part, corresponding to the lower case notation for the non-seasonal part.

\subsection{The FFORMA Algorithm}
\label{appendix:base}

FFORMA is a hybrid algorithm for time series forecast proposed by \citet{FFORMA2018}. Based on XGBoost, a gradient tree boosting model from \citet{Chen2016}, FFORMA combines a set of naive forecasting methods, such as the MA, ETS and ARIMA, to improve the forecasting accuracy. The training procedure is given in Table \ref{table:FFORMA}. Afterwards, based on the trained XGBoost model, the predictions are given by Eq.~(\ref{eq:FFORMA_prediction}) in Section \ref{subsec:practical}.
\begin{table}[!htbp]
\caption{Pseudocode for FFORMA Training \citep{FFORMA2018}}
\small
\centering
{\def\arraystretch{0.8}
\begin{tabular}{lll }
\hline
\hline
 \textbf{Input:} \\
\ \ \ \ $\{\bm{y}_1,\cdots,\bm{y}_N\}$: $N$ observed time series;\\
\ \ \ \ $F$: Set of functions for calculating features;\\
\ \ \ \ $M$: Set of naive forecasting methods.\\
\hline 
\textit{Data Preparation}\\
For $n=1$ to $N$: \\
\ \ \ \ (a) Split $\bm{y}_n$ into a training and test periods(sets);\\
\ \ \ \ (b) Calculate the set of features $f_n\in F$ over the training set;\\
\ \ \ \ (c) Fit every forecasting method $m\in M$ over the training set, and generate forecasts over the test set;\\
\ \ \ \ (d) Calculate forecast losses $L_m(f_n)$ with feature $f_n$ and method $m$ over the test set.\\
\textit{Model Training}\\
\ \ \ \ Train an XGBoost model on $\bm{w}=\{w_{m}, \forall m\in M\}$ by minimizing $\textit{Loss}=\sum_{n} \sum_{m}  w_{m}(f_n)\times L_{m}(f_n)$.\\
\hline
\hline
\end{tabular}
\label{table:FFORMA}
}
\end{table}

\subsection{GBDT and XGBoost}
\label{appendix:GBDT}
Proposed by \citet{Friedman2001}, GBDT is a tree-based numerical heuristic. It consists of a series of sub-trees, in which each tree fits the residual between actual values and current predictions from previous trees. The final output is the addition of the outputs from subtrees. Specifically, let
$\bm{x}$ denote the input features, $\bm{y}=\{y_i, \ \forall i\}$ denote actual target values, and set $F_0(\bm{x})=\text{average}(\bm{y})$, then the calculation procedure is an iterative process of
\begin{enumerate}
\item $\tilde{y}_i=y_i-F_{n-1}(\bm{x}_i)$, $\forall i$;
\item $(\rho_n,\bm{a}_n)=\argmin_{\rho,\bm{a}}\sum_i \left(\tilde{y}_i-\rho h(\bm{x}_i;\bm{a})\right)^2$;
\item $F_{n}(\bm{x})=F_{n-1}(\bm{x})+\rho_n h(\bm{x};\bm{a})$;
\end{enumerate}
until the termination condition is met, where $F_i(\bm{x})$is the predictions after the $m$th iteration, and $\rho$,$h(\bm{x};\bm{a}_m)$are the parameters for the new sub-tree in the $m$th iteration.

The XGBoost algorithm is a specific case of GBDT. Specifically, when generating the sub-trees, XGBoost uses the Taylor expansion with first and second gradients to approximate the loss function and further incorporate a regularization term penalizing the model complexity in the objective function. As a result, compared with classical GBDT, XGBoost shows better performance in accuracy and efficiency.
We refer the readers to \citet{Chen2016} for the fundamental mathematics
and the operational details if interested.

\subsection{The STL Algorithm}
\label{appendix:stl}

As mentioned before, the STL algorithm is a classical decomposition algorithm proposed by \citet{Cleveland1990}, and the detailed pseudocode is given in Table \ref{table:STL}. In our case, we omit the component of residuals $\bm{l_R}$, and the forecasted value $\bm{l}=\bm{l_T}+\bm{l_S}$, 
where $\bm{l_T}$ and $\bm{l_S}$ are predicted by corresponding Loess algorithms.

\begin{table}[!htbp]
\caption{Pseudocode for STL Decomposition \citep{Cleveland1990}}
\small
\centering
{\def\arraystretch{0.8}
\begin{tabular}{lll }
\hline
\hline
 \textbf{Input:} time series data $\bm{l}$\\
 \textbf{Parameters:} \\
\ \ \ \ $n_{(p)}$: number of samples for a seasonality or period;\\
\ \ \ \ $n_{(s)}$, $n_{(t)}$, $n_{(l)}$: parameters for corresponding Loess.\\
\textbf{Definition:} Subseries: the series within a period.\\
\hline 
\textit{0. Initialization:} $\bm{l_T}^{(0)}=0$, $\bm{\delta}^{(0)}=1$.\\
For $k= 0,1,\cdots, K$:\\
\textit{1. Inner loop:}\\
\ \ \ \ (a) Detrending: Calculate $\bm{l}-\bm{l_T}^{(k)}$.\\
\ \ \ \ (b) Cycle-subseries smoothing: Use Loess with $q=n_{(s)}$, $d=1$,  $\delta=\bm{\delta}^{(k)}$ to regress the subseries; \\
\ \ \ \ \ \ \ \ \ Get the temporary seasonal series $\bm{C}^{(k+1)}$, which is of length $N+2n_{(p)}$ (from $-n_{(p)}+1$ to $N+n_{(p)}$).\\
\ \ \ \ (c) Low-pass filtering: \\
\ \ \ \ \ \ \ \ \ (i) Calculate the moving average of $n_{(p)}$  on  $\bm{C}^{(k+1)}$;\\
\ \ \ \ \ \ \ \ \ (ii) Calculate the moving average of $3$ on  the previous result from (i);\\
\ \ \ \ \ \ \ \ \ (iii) Use Loess with $q=n_{(l)}$ and $d=1$ to regress the previous result from (ii);\\
\ \ \ \ \ \ \ \ \ (iv) Get the low-pass result $\bm{L}^{(k+1)}$, which is of size $N$ (from time $1$ to $N$).\\
\ \ \ \ (d) Detrending of smoothed cycle-subseries: Update $\bm{l_S}^{(k+1)}=\bm{C}^{(k+1)}-\bm{L}^{(k+1)}.$\\
\ \ \ \ (e) Deseasonalizing: Calculate $\bm{l}-\bm{l_S}^{(k+1)}$.\\
\ \ \ \ (f) Trend smoothing: \\
\ \ \ \ \ \ \ \ \  Use Loess with $q=n_{(t)}$, $d=1$ and $\delta=\bm{\delta}^{(k)}$ to regress the previous result from (e), and get $\bm{l_T}^{(k+1)}$.\\
\textit{2. Outer loop:}\\

\ \ \ \ Calculate $\bm{l_R}^{(k+1)}=\bm{l}-\bm{l_T}^{(k+1)}-\bm{l_S}^{(k+1)}$;\\
\ \ \ \  Calculate robustness weight $\bm{\delta}^{(k+1)}=B\left(\frac{|\bm{l}|}{6 median (|\bm{l}|)}\right)$, where  $B(u)=\left\{ \begin{array}{ll}(1-u^2)^2 &\textit{ if } 0\leq u<1, \\ 0  &\textit{ if } u\geq 1.\end{array} \right. $.\\
\textit{3. Repeat step 1 and 2 until $k==K$ or $|\bm{l_T}^{(k)}-\bm{l_T}^{(k-1)}|+|\bm{l_S}^{(k)}-\bm{l_S}^{(k-1)}|\leq \varepsilon$}.\\

\hline
\hline
\end{tabular}
\label{table:STL}
}
\end{table}

\newpage
\section{Supplementary Materials for Numerical Experiments}
\label{sec:supplement_results}
\vspace{0.2cm}
\subsection{Supplementary Details of Experiment Settings}
\label{subsec:features}
\begin{table}[!htbp]
\vspace{-0.2cm}
\caption{Features in the Practical Case of JD.com's Sales Forecast}
\small
\centering
{\def\arraystretch{0.8}
\begin{tabular}{lll }
\hline
\hline
 \textbf{Stage}& \textbf{Model} & \textbf{Features} \\
\hline
\multirow{3}{4cm}{\makecell[l]{\textbf{Stage 1: }Preliminary\\ decomposition}}
  &Baseline       & Historical sales \\
\cline{2-3}
  &Promotion       & \makecell[l]{X: Promotion types in sales history\\[-0.2cm] Y: Historical sales \\[-0.2cm] T: Prices in sales history\\[-0.2cm]W: Reference price, year, month, day, weekday}  \\
\cline{2-3}
 &Festival      & \makecell[l]{Historical sales\\[-0.2cm]  Festival level in sales history} \\
\hline
\makecell[l]{\textbf{Stage 2: }ML improvement\\[-0.2cm] based on MQ-CNN} &- &\makecell[l]{Historical Sales\\[-0.2cm]  Time features: Year, month, day, weekday \\[-0.2cm] Product features: length of sales, product id, category, brand\\[-0.2cm] $l$: Decomposed predictions from stage 1}\\
\hline
\hline
\end{tabular}

\label{table:practical_features}
}
\end{table}

\begin{table}[!htbp]
\vspace{-0.4cm}
\caption{Parameters in Numerical Experiments}
\footnotesize
\centering
{\def\arraystretch{0.8}
\begin{tabular}{ll l l}
\hline
\hline
 \textbf{Parameter} & \textbf{Description}& \textbf{Value} & \textbf{Experiment} \\
\hline
\multirow{2}{2cm}{$T$} & \multirow{2}{6.5cm}{Length of historical data} &72 &Practical\\
&  & 60&Public \\
\multirow{2}{2cm}{$H$} & \multirow{2}{6.5cm}{Forecasting horizon (days)}&24&Practical\\
& &30 or 31 &Public\\
\multirow{2}{2cm}{$N$} & \multirow{2}{6.5cm}{Number of preliminary decompositions}&3 &Practical\\
 & &2 &Public\\
Learning rate& Learning rate in stage 2 &1e-3  &All\\
\multirow{2}{2cm}{Batch size} & \multirow{2}{7cm}{\makecell[l]{Number of samples in each batch, include \\[-0.2cm] the features of 32 products, each with $T$ samples }} &2304 (32*72) &Practical \\
  & &1920 (32*60)&Public \\
Number of batches (by sku) & Number of batches for each epoch&100&All \\
Epoch & Number of iterations in training &10-50&All\\
Repetition time& Number of repetitions of experiments &10  &All\\
Units in the cell of Wavenet& Parameter of Wavenet& 32&All\\
Number of hidden layers in Wavenet& Parameter of Wavenet& 3&All\\
Units in  MLP& Parameter of MLP& 32&All\\
$n_{(p)}$ & Seasonality parameter in STL &12 &Public \\
$n_{(s)},n_{(t)},n_{(l)}$& Parameter for Loess in STL & 1 &Public \\
\hline
\hline
\end{tabular}
\label{table:parameters}
}
\end{table}

\newpage 
\subsection{Python Packages in Implementation}
\label{appendix:packages}

\begin{table}[!htbp]
\caption{Python Packages in Implementation}
\scriptsize
\centering
{\def\arraystretch{1.5}
\begin{tabular}{lll l}
\hline
\hline
 \textbf{Name} &\textbf{Description}&  \textbf{Reference Link}\\
\hline
statsmodels & Holt-Winters \& ARIMA  & \url{https://pypi.org/project/statsmodels/}\\
EconML & DML   & \url{https://github.com/microsoft/EconML}\\
fforma & FFORMA   & \url{https://github.com/christophmark/fforma}\\
STLDecompose & STL & \url{https://github.com/jrmontag/STLDecompose}\\
gluonts.model.wavenet & Wavenet  & \url{https://ts.gluon.ai/api/gluonts/gluonts.model.wavenet.html}\\
gluonts.model.seq2seq & MQ-CNN   & \url{https://ts.gluon.ai/api/gluonts/gluonts.model.seq2seq.html}\\
gluonts.model.deepar & DeepAR  & \url{https://ts.gluon.ai/api/gluonts/gluonts.model.deepar.html}\\
gluonts.model.n\underline{ }beats & N-BEATS & \url{https://ts.gluon.ai/api/gluonts/gluonts.model.n_beats.html}\\
\hline
\hline
\end{tabular}
\label{table:python_packages}
}
\end{table}

\subsection{Supplementary Numerical Results}
\label{appendix:practice_detail}

\begin{table}[htbp]
\centering
\captionof{table}{Comparison on Model Performance by Month (Practical Data)}
\label{table:practice_detail}
\footnotesize
{\def\arraystretch{0.8}
\begin{tabular}{c lcc cl cc}
\hline
\hline
\textbf{Month}  &\textbf{Algorithm}                     &\textbf{RMSE}        & \textbf{P50\underline{ }QL}     & \textbf{Month}  &\textbf{Algorithm}                     &\textbf{RMSE}        & \textbf{P50\underline{ }QL}  \\
\hline
Jun. & MQ-CNN                         & 3.1469&0.2946 & Aug.& MQ-CNN                         & 1.9988&0.1989 \\
Jun. &\makecell[l]{Preliminary Decomposition \uppercase\expandafter{\romannumeral1}\\[-0.2cm] \textit{(Additive Combination)}}    &  2.9050& 0.2452&Aug. &\makecell[l]{Preliminary Decomposition \uppercase\expandafter{\romannumeral1}\\[-0.2cm]  \textit{(Additive Combination)}} &1.9799& 0.1960\\
Jun. &\makecell[l]{Preliminary Decomposition \uppercase\expandafter{\romannumeral2}\\[-0.2cm]  \textit{(Simple MLP Combination)}}    &  3.0391  &  0.2829& Aug. &\makecell[l]{Preliminary Decomposition \uppercase\expandafter{\romannumeral2}\\[-0.2cm]  \textit{(Simple MLP Combination)}}&1.9205&0.1885\\
Jun. &W-R Algorithm ($\alpha=1$)          & 2.2872 &0.2213 &Aug. &W-R Algorithm ($\alpha=1$)  &1.8987  &0.1830\\
Jun. &DeepAR                        &  3.3236   &   0.3281&Aug. &DeepAR                        &    2.3000 & 0.2648 \\
Jun. &N-BEATS                        & 3.4232 &  0.3632 &Aug. &N-BEATS                        & 2.3143 &  0.2776     \\
Jul. & MQ-CNN                         & 2.4616& 0.2106&Sep. & MQ-CNN                         &2.4356 &   0.2255\\
Jul. &\makecell[l]{Preliminary Decomposition \uppercase\expandafter{\romannumeral1}\\[-0.2cm]  \textit{(Additive Combination)}}    & 2.4021 & 0.2134&Sep. &\makecell[l]{Preliminary Decomposition \uppercase\expandafter{\romannumeral1}\\[-0.2cm]  \textit{(Additive Combination)}} &2.3547&0.2227 \\
Jul. &\makecell[l]{Preliminary Decomposition \uppercase\expandafter{\romannumeral2}\\[-0.2cm]  \textit{(Simple MLP Combination)}}    &  2.3633  & 0.2015 &Sep. &\makecell[l]{Preliminary Decomposition \uppercase\expandafter{\romannumeral2}\\[-0.2cm]  \textit{(Simple MLP Combination)}}&2.1745&0.2111\\
Jul. &W-R Algorithm ($\alpha=1$)          &1.8987  &0.1954 &Sep. &W-R Algorithm ($\alpha=1$)  &2.0738 &0.2009 \\
Jul. &DeepAR                        &   2.4727  &0.2252   &Sep. &DeepAR                        &  2.4374   & 0.2260\\
Jul. &N-BEATS                        & 2.5023 & 0.2410  &Sep. &N-BEATS                        & 2.7691 &   0.3250    \\
\hline
\hline  
\end{tabular}
}
\end{table}

\end{document}